\documentclass[10pt,conf,onecolumn,compsoc]{IEEEtran}
\ifCLASSOPTIONcompsoc
  \usepackage[nocompress]{cite}
\else
  \usepackage{cite}
\fi
\ifCLASSINFOpdf
\else
\fi
\newcommand{\etal}{\emph{et al.}}
\usepackage{booktabs}
\usepackage{multirow}
\usepackage{graphicx}
\usepackage{amsmath}
\usepackage{amsfonts}
\usepackage{enumitem}

\usepackage{makeidx}
\usepackage{color}
\usepackage{tikz}
\usepackage{subfig}
\usepackage{graphics}
\usepackage[export]{adjustbox}
\usepackage{todonotes}

\usepackage{algorithm}
\usepackage{algpseudocode}

\usepackage[usestackEOL]{stackengine}
\usepackage[normalem]{ulem}

\usepackage{textcomp}

\begin{document}
%\title{GANuscripting/GANscript/GANlligraphy: Generation of Handwritten Text-line Image with Content and Calligraphy Conditioning}
%\title{GANuscripting: Generation of Handwritten Text-line Images with Content and Calligraphy Conditionning \textcolor{magenta}{for HTR???????}}

% random option 1
%\title{GANuscripting: An effective Boost on Handwritten Text-line Recognition with Generated Samples Conditioned on Visual Appearance and Textual Content}

% random option 2..
\title{Content and Style Aware Generation of Text-line Images for Handwriting Recognition}

% random option 3..

% random option 4..

% random option 5..

% GANscript

\author{Lei~Kang,
        Pau~Riba,
        Mar\c{c}al~Rusi{\~n}ol,
        Alicia~Forn\'{e}s,
        Mauricio~Villegas% <-this % stops a space
\IEEEcompsocitemizethanks{\IEEEcompsocthanksitem L. Kang is with Computer Science Dept., Shantou University, China. \hfil\break
E-mail: lkang@stu.edu.cn
\IEEEcompsocthanksitem P. Riba is with Helsing AI, Munich, Germany. \hfil\break
E-mail: pau.riba@helsing.ai
\IEEEcompsocthanksitem M. Rusi{\~n}ol is with AllRead MLT, Barcelona, Spain. \hfil\break
E-mail: marcal@allread.ai
\IEEEcompsocthanksitem A. Forn\'{e}s is with Computer Vision Center, Computer Science Dept., Universitat Autonoma de Barcelona, Spain.\protect \hfil\break
E-mails: afornes@cvc.uab.es
\IEEEcompsocthanksitem M. Villegas is with omni:us, Berlin, Germany. \hfil\break
E-mail: mauricio@omnius.com}
\thanks{Manuscript version, accepted to TPAMI}}

\markboth{Journal of \LaTeX\ Class Files,~Vol.~14, No.~8, August~2015}%
{Shell \MakeLowercase{\textit{et al.}}: Bare Demo of IEEEtran.cls for Computer Society Journals}



\IEEEtitleabstractindextext{%
\begin{abstract}
Handwritten Text Recognition has achieved an impressive performance in public benchmarks. However, due to the high inter- and intra-class variability between handwriting styles, such recognizers need to be trained using huge volumes of manually labeled training data. To alleviate this labor-consuming problem, synthetic data produced with TrueType fonts has been often used in the training loop to gain volume and augment the handwriting style variability. However, there is a significant style bias between synthetic and real data which hinders the improvement of recognition performance. To deal with such limitations, we propose a generative method for handwritten text-line images, which is conditioned on both visual appearance and textual content. Our method is able to produce long text-line samples with diverse handwriting styles. Once properly trained, our method can also be adapted to new target data by only accessing unlabeled text-line images to mimic handwritten styles and produce images with any textual content. Extensive experiments have been done on making use of the generated samples to boost Handwritten Text Recognition performance. Both qualitative and quantitative results demonstrate that the proposed approach outperforms the current state of the art. 
\end{abstract}

}

\maketitle

\IEEEdisplaynontitleabstractindextext

% \fi
\IEEEpeerreviewmaketitle

%%%%%%%%%%%%%%%%%%%%%%%%%%%%%%%%%%%%%%%%%%%%%%%%%%%%%%%%%%%%%%%%%%%%%%
%%%%%%%%%%%%%%%%%%%%%%%%%%%%%%%%%%%%%%%%%%%%%%%%%%%%%%%%%%%%%%%%%%%%%%
\IEEEraisesectionheading{\section{Introduction}\label{sec:introduction}}

\label{sec:intro}

\IEEEPARstart{D}{ocument} analysis and recognition is challenging because of the varied document types, ranging from  historical documents to modern administrative ones. In the case of documents containing handwritten text, the inter- and intra- writer variability of handwriting styles hinder the recognition performance of Handwritten Text Recognition (HTR) methods. Since manually labeling lots of real handwritten text images is labor-consuming, the utilization of data augmentation and synthetic data generation using TrueType fonts is a common practice to boost the HTR performance~\cite{krishnan2019hwnet}. However, the style bias between the synthetic and real data hinders the improvement of the recognition performance. 

Since Generative Adversarial Networks (GANs)~\cite{goodfellow2014generative} were firstly introduced in 2014, we have witnessed a remarkable success in generating natural scene images, which are even indiscernible from real ones by humans~\cite{karras2020analyzing}. Conditional Generative Adversarial Networks (cGANs)~\cite{mirza2014conditional} were proposed to condition the generation process with a class label. Thus, controllable samples can be generated from different given types~\cite{choi2018stargan}. However, these conditioned class labels have to be predefined and hard-coded in the model before the training process, so that it lacks the flexibility to generate images from unseen classes at inference time.

Concerning the specific case of generating samples of handwritten text, there are two different approaches to the problem. Since handwritten text is a sequential signal in nature, the same as natural language strings~\cite{yu2017seqgan}, sketch drawings~\cite{ha2017neural,zheng2019strokenet}, audio signals~\cite{dong2018musegan} or video streams~\cite{tulyakov2018mocogan}, it is natural that the first attempts at generating handwritten data~\cite{graves2013generating} were based on Recurrent Neural Networks (RNNs)~\cite{lipton2015critical}. Such approaches generate a sequence of strokes in vectorial format that are used to render images. On the contrary, some more recent approaches propose to directly generate images instead of sequences of strokes. By producing images directly, long-range dependency and gradient vanishing problems of recurrencies are avoided, while achieving a better efficiency. Furthermore, such approaches are able to produce richer results in the sense that they go beyond producing just nib locations, but also provide visual appearance such as the calligraphic styles, such as slant, glyph shapes, stroke width, darkness, character roundness, ligatures, etc., and background paper features like texture, opacity, show-through effects, etc.

\begin{figure*}[t!]
    \centering
    %\resizebox{\linewidth}{!}{
    %\hspace*{-2.5cm}
    \begin{tabular}{l}
    \toprule
    \includegraphics[height=0.6cm,valign=c]{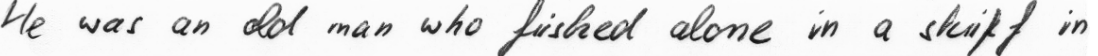}\\
    \includegraphics[height=0.6cm,valign=c]{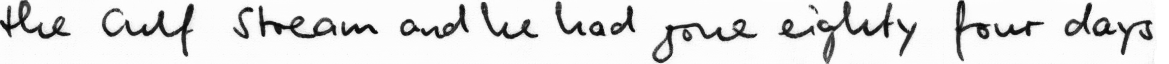}\\
    \includegraphics[height=0.6cm,valign=c]{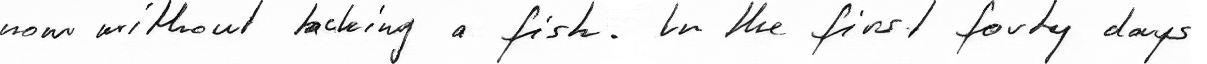}\\
    \includegraphics[height=0.6cm,valign=c]{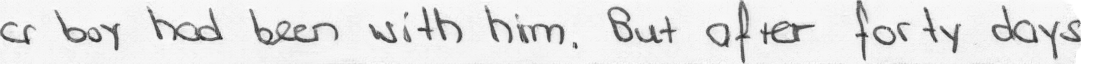}\\
    \includegraphics[height=0.6cm,valign=c]{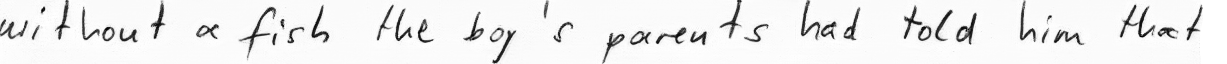}\\
    \includegraphics[height=0.6cm,valign=c]{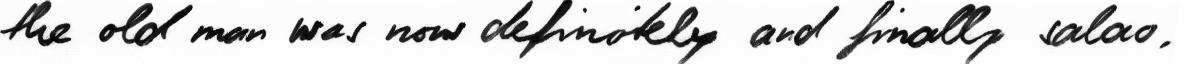}\\
    \includegraphics[height=0.6cm,valign=c]{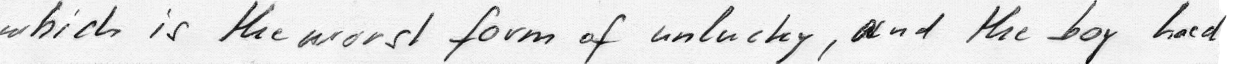}\\
    \includegraphics[height=0.6cm,valign=c]{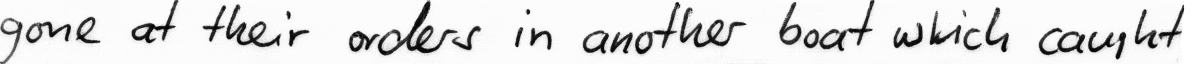}\\
    \includegraphics[height=0.6cm,valign=c]{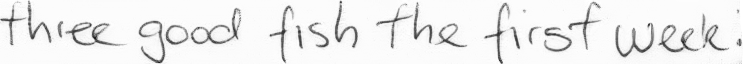}\\
    \bottomrule
\end{tabular}
%\hspace*{-2.5cm}
%}%
    \caption{Examples of generated text-lines, each one using a different handwriting style. The text corresponds to the first paragraph of the book ``The Old Man and the Sea''.}
    \label{fig:old_man}
\end{figure*}

Current state-of-the-art methods that directly generate handwriting images work at different levels. First, some approaches are focused on producing isolated characters or ideograms~\cite{gregor2015draw,chang2018generating}. Such approaches often work over a set of predefined classes, so that they can only generate a reduced set of contents. Second, some approaches are able to generate handwritten words~\cite{alonso2019adversarial,kang2020ganwriting}, allowing not to be restricted to a closed vocabulary. Finally, some works like~\cite{fogel2020scrabblegan,davis2020text} go beyond isolated words and produce full text-lines. The generation on text-line level is difficult because not only the handwritten text should be readable and realistic, but also the writing flow should be natural and smooth. 

Since we aim to boost the HTR performance at text-line level, in this work we propose a method for generating handwritten text-line images. By conditioning on both calligraphic style from handwritten images and textual content from an external text corpus, our proposed method is able to produce realistic, writer agnostic and readable samples for handwritten text-lines (see Figure~\ref{fig:old_man}), which can be effectively used in order to train and improve the final HTR performance.

This work supposes a significantly extended version of our previous conference paper~\cite{kang2020ganwriting}. In this work, we have enhanced our previous generative architecture in order to generate whole sentences rather than single words, where Periodic Padding and Transformer-based Recognizer are newly proposed. In addition, we propose a novel version of the Fr\'{e}chet Inception Distance (FID) metric to guide the method to choose the best hyper-parameters specifically for variable-length samples like handwritten text images. In the training step, we make use of curriculum learning strategy to help the proposed method to  generalize from short text-lines to longer ones. More importantly, and contrary to our previous work (in which the only goal was to generate realistic text images), the proposed method is particularly focused on improving the HTR performance, demonstrating that the use of realistic synthetic generated text at training time is indeed useful for improving HTR. 

To summarize, the main contributions of this paper are the following:
\begin{itemize}
    \item We propose a novel method for handwritten text-line image generation conditioning on textual content and visual appearance information, which is capable of generating open vocabulary text and visual appearance.
    \item We introduce an improved version of the FID measure, namely vFID, as a novel metric to evaluate the quality of the generated handwritten image. It is more robust to variable-length images and particularly suited for the handwriting case.
    \item We conduct extensive experiments to demonstrate, on the one hand, the realism of the generated handwritten text images and, on the other hand, the boost in HTR performance avoiding the manual labeling effort.
\end{itemize}
The rest of the paper is organized as follows. In Section~\ref{sec:relate} we introduce the state-of-the-art approaches related to handwriting generation. In Section~\ref{sec:method} we explain our proposed method in details with different modules. In Section~\ref{sec:vfid} the proposed novel vFID metric is introduced. In Section~\ref{sec:exp} extensive qualitative and quantitative experiments are presented and discussed. Finally, Section~\ref{sec:con} draws the conclusions of this work.

%%%%%%%%%%%%%%%%%%%%%%%%%%%%%%%%%%%%%%%%%%%%%%%%%%%%%%%%%%%%%%%%%%%%%%
%%%%%%%%%%%%%%%%%%%%%%%%%%%%%%%%%%%%%%%%%%%%%%%%%%%%%%%%%%%%%%%%%%%%%%
\section{Related Work}
\label{sec:relate}

Traditional methods~\cite{wang2005combining,lin2007style,konidaris2007keyword,thomas2009synthetic} approached the generation of word samples by manually segmenting individual characters or glyphs and then tune a deformation to match the target writing style. Recently, based on these rendering methods, Haines~\etal~\cite{haines2016my} succeeded in generating indistinguishable historical manuscripts of Sir Arthur Conan Doyle, Abraham Lincoln and Frida Kahlo with new textual contents, but these impressive results are obtained at the cost of a high manual intervention.

Editing text in the natural scene images aims to replace a word in the source image with a new one while maintaining the original style. Wu~\etal~\cite{wu2019editing} proposed an end-to-end trainable style retention network (SRNet) for the text editing task, which is the first work to edit text image in the word-level. Roy~\etal~\cite{roy2020stefann} developed both Font Adaptive Neural Network (FANnet) and Colornet architectures to modify text in an natural scene image at character-level. Yang~\etal~\cite{yang2020swaptext} also succeeded in manipulating the texts from the natural scene images even with severe geometric distortion. However, the texts in the natural scene images are often to be typed fonts, especially in the datasets evaluated with these methods. The lack of cursive styles of the scene texts makes these approaches hard to work in handwriting scenarios.

The generation of sequential handwritten data consists of producing stroke sequences in vector form with nib locations and sometimes velocity records. With the coming of deep learning era, Graves~\cite{graves2013generating} utilized Long Short-Term Memory (LSTM) to predict point by point at each time step to generate stroke sequences conditioned on a given writing style and a certain text string. Zhang~\etal~\cite{zhang2017drawing} investigated RNN as both discriminative and generative models for recognizing and drawing cursive handwritten Chinese characters. Following this sequential-based idea, some recent works~\cite{ha2017neural,ganin2018synthesizing,zheng2019strokenet,mayr2020spatio} have reached an impressive performance on text or sketch generation. Online handwritten data preserves rich dynamic information such as trajectory, velocity and pressure, which is a big advantage over the offline data. However, the offline data maintains richer visual appearance information such as stroke thickness, ink shading and paper textures. In this paper, we focus on the problems on generating realistic handwritten images at pixel-level, so we only make use of offline handwritten data.

Different levels of offline handwritten data can be processed: characters/glyphs, words and text-lines. Based on the ideas of variational auto-encoders~\cite{kingma2013auto} or GANs~\cite{goodfellow2014generative}, some works achieve impressive performance on synthesizing Chinese ideograms~\cite{lyu2017auto,tian2017zi2zi,chang2018generating,jiang2018w,wu2020calligan} and glyphs~\cite{azadi2018multi}. However, these methods are restricted to a predefined set of content classes and the input images have fixed size. To overcome the limitation of incapability of generating out of vocabulary (OOV) texts, Alonso \emph{et al.}~\cite{alonso2019adversarial} proposed a cGAN-based method to generate handwritten word samples, which is conditioned on RNN-embedded text information. However, this proposed approach suffers from the mode collapse problem so that it learns the general writing style of the training set and does not offer variability of the generated samples. Our previous work, GANwriting~\cite{kang2020ganwriting}, works only at word level, so that it cannot render long text strings into a handwritten image. To keep the consistency in a text-line image, the extension from word to text-line level is needed. Fogel~\emph{et al.}~\cite{fogel2020scrabblegan} equip a style-promoting discriminator to be able to generate diverse styles for handwritten image samples. However, the generated characters have the same receptive field width, which can make the generated samples look unrealistic. Davis~\etal~\cite{davis2020text} takes advantage of CTC activations~\cite{graves2006connectionist} to produce spaced text, which helps the generator to achieve horizontal alignment with the input style image. The style information is the concatenation of both global style feature and character-wise style feature. However, the character-wise style feature highly depends on the performance of CTC, so mode collapse problem may happen when tackling the unseen style images from the target dataset.  

In summary, all the above described methods are not robust enough to produce high quality handwritten samples with a huge diversity in handwriting styles, especially when producing longer text-lines. In Section~\ref{sec:method}, we describe our generative method for handwritten text-line images with carefully designed modules.

%%%%%%%%%%%%%%%%%%%%%%%%%%%%%%%%%%%%%%%%%%%%%%%%%%%%%%%%%%%%%%%%%%%%%%
%%%%%%%%%%%%%%%%%%%%%%%%%%%%%%%%%%%%%%%%%%%%%%%%%%%%%%%%%%%%%%%%%%%%%%

\begin{figure*}[ht!]
    \centering
    \includegraphics[width=\linewidth]{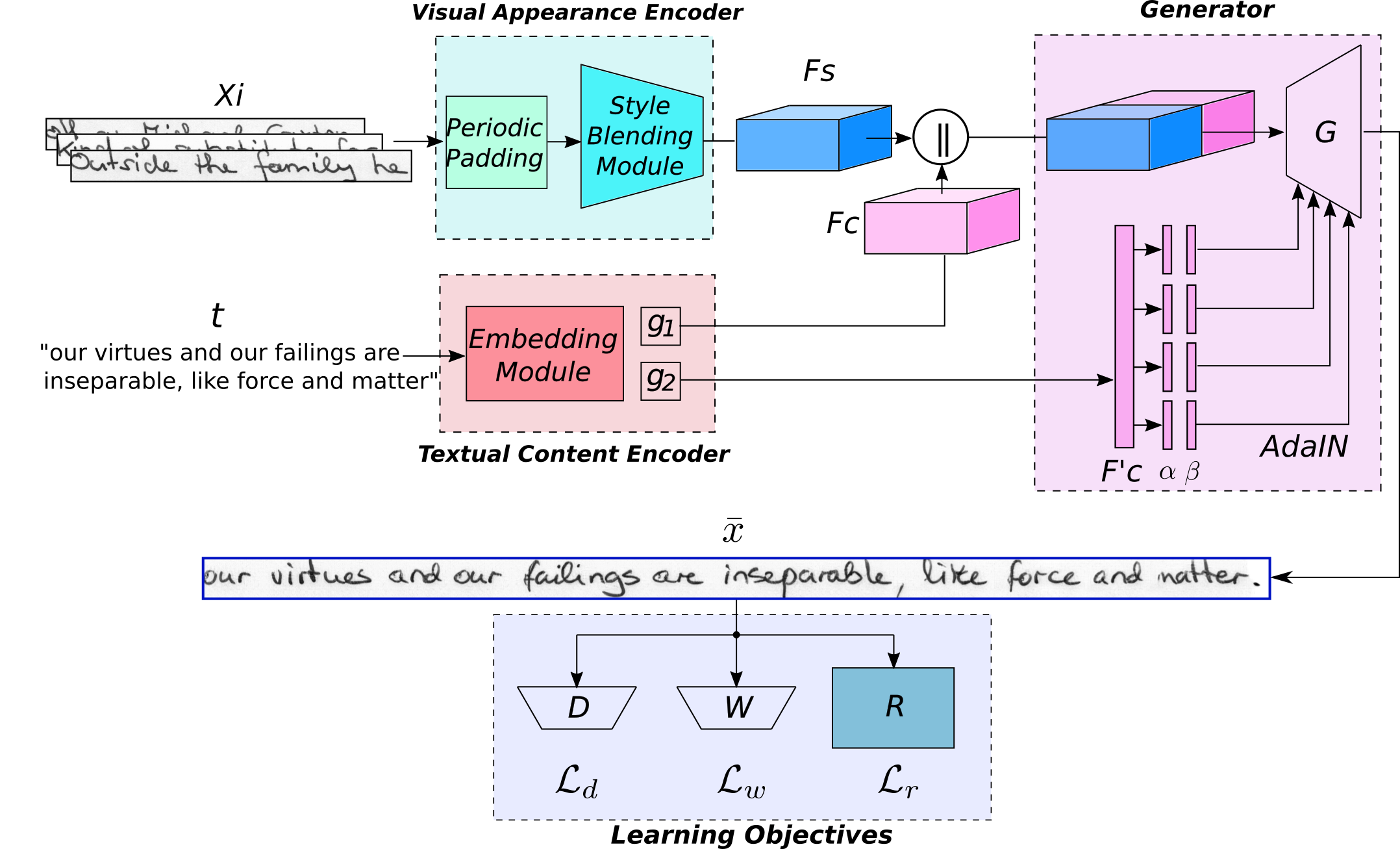}
    \caption{Architecture of the proposed handwriting synthesis model. It consists of a Visual Appearance Encoder (cyan box), a Textual Content Encoder (red box), a Generator (magenta box) and learning objectives (blue box). $X_i$ and $t$ are the images and text string input, respectively. The $\bar{x}$ is the generated sample that shares the visual appearance with $X_i$ and contains the textual information with $t$.}
    \label{fig:arch}
\end{figure*}

\section{Handwritten Text-line Synthesizer}
\label{sec:method}

\subsection{Problem Formulation}

Let $\{\mathcal{X},\mathcal{Y},\mathcal{W}\}=\{(x_i, y_i, w_i)\}_{i=1}^N$ be a multi-writer handwritten text-line dataset, containing gray-scale text-line images $\mathcal{X}$, their corresponding transcription strings $\mathcal{Y}$ and their writer identifiers $\mathcal{W}$. In this work, the handwriting calligraphic style is considered as an inherent feature for each of the different writers, and we also hypothesize that the background paper features are consistent within each writer. Thus, the visual appearance is identified with $w_i\in\mathcal{W}$. Therefore, let $X_i = \{x_{w_i,j}\}_{j=1}^K \subset \mathcal{X}$ be a subset of $K$ real text-line images with the same style defined by writer $w_i\in\mathcal{W}$. Besides, $\mathcal{A}$ denotes the alphabet containing all the supported characters such as lower and upper case letters, digits and punctuation signs that the generator will be able to produce.

In this setting, the realistic handwritten text-line generation problem is formulated in terms of few-shot learning. Two inputs are given to the model: 1) a set of images $X_i$ as a support example of the visual appearance attributes of a particular writer $w_i$; and 2) a textual content provided by any text string $t$ where $t_n\in\mathcal{A}$. The proposed conditioned handwritten text generation model is able to combine both sources of information in order to yield realistic handwritten text-line images, which share the visual appearance attributes of writer $w_i$ and the textual content provided by the string $t$. Finally, our objective model $H$, able to generate handwritten text, is formally defined as
\begin{equation}
    \bar{x} = H\left(t, X_i\right) = H\left(t, \left\{x_1, \ldots, x_K\right\}\right),
\end{equation}
where $\bar{x}$ is the artificially generated handwritten text-line image with the desired properties. From now on, we denote $\bar{\mathcal{X}}$ as the output distribution of the generative network $H$.

Figure~\ref{fig:arch} shows a detailed overview of the proposed architecture. The proposed model consists of four main components: the Visual Appearance Encoder, the Textual Content Encoder, the Generator and the learning objectives. On the one hand, the generator, which is conditioned by a combination of visual appearance attributes and textual content information, is able to produce human-readable handwritten text-line images. On the other hand, three learning objectives are proposed to guide the learning process towards generating realistic images, which are classified within a particular visual appearance while sharing the specified textual content.

\subsection{Visual Appearance Encoder}

The visual appearance encoder receives as input a given set $X_i$ of handwritten text line images from a particular writer $w_i$. We assume that these given images share some visual appearance features that are inherent to each writer. These visual appearance attributes consist of properties such as slant, glyph shapes, stroke width, character roundness, ligatures, etc. In our proposed approach, the visual appearance encoder aims at extracting the handwriting style attributes from the set of images $X_i$. With this aim, the textual content information from those images is ignored and totally disentangled from the stylistic visual attributes.

The proposed visual appearance encoder consists of two modules: first, a periodic padding module which ensures that all the images share the same size and, second, a style blending module in charge of extracting the visual appearance features. In general, the visual appearance encoding process is denoted as $F_s = S\left(X_i\right)$.

\subsubsection{Periodic Padding Module}

\begin{figure*}[ht!]
    \centering
    \begin{tabular}{l}
    \toprule
    \includegraphics[height=0.4cm]{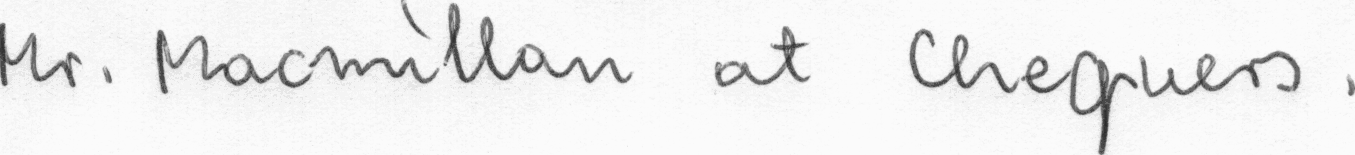}\\
    \includegraphics[height=0.4cm]{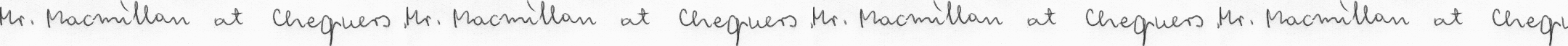}\\
    \midrule
    \includegraphics[height=0.4cm]{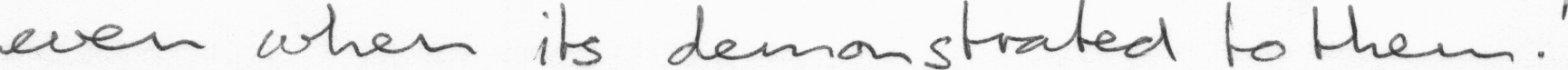}\\
    \includegraphics[height=0.4cm]{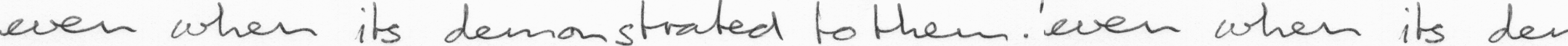}\\
    \bottomrule
    \end{tabular}
    \caption{Periodic padding example. Given a real image, periodic padding to the right is applied several times until the maximum image width $L$ is reached.}
    \label{fig:filling}
\end{figure*}

As the style image samples $X_i$ have varied shapes, they are firstly resized to the same $64$ pixels height while keeping the aspect ratio. Let $L$ be the maximum length of both input and output images. To mimic the background color with our padding, the input image $x_j \in X_i$ is first normalized within the range $[0,1]$ and then their intensities are inverted $1-I/255$. Thus, the writing strokes have values close to $1$ whereas the background has values close to $0$. In the HTR literature, the usual technique to align all the images to have the same length in a mini-batch is to add $0$-padding to the right of each image until reaching the maximum length $L$. We have experimentally observed that $0$-padding has a severe impact on the handwritten text-line image generation process, which can easily collapse in terms of style in the padded regions. This is especially important when there is a huge difference in the length of input images. The style representations $F_s$ contain not only the visual appearance attributes, but also the spatial information. Thus, the padding would make longer texts to loose the handwritten style consistency in the generated output. To overcome this problem, we introduce a simple periodic padding module, which consists in repeating the input image several times to the right until the length fits the maximum width $L$. An example is shown in Figure~\ref{fig:filling}. Thus, the periodic padding can deal with the visual appearance vanishing problem, which is especially useful to generate long text handwritten samples with short input images. Bear in mind that the style images $X_i$ are only used to extract style features, which are completely independent from the textual content in the image.

\subsubsection{Style Blending Module}

The $K$ images from $X_i$, now all having the same size are channel-wise concatenated and given as input for the subsequent stylistic feature extractor. The style blending module is a sequence of convolutional layers. It is in charge of producing the visual appearance features $F_s$ from the set of images $X_i$, which is represented as a $64\times L\times K$ input tensor. The choice of the convolutional architecture is detailed in Section~\ref{sec:exp}.

%Style blending module is a sequence of Convolutional Neural Network (CNN) layers to extract visual appearance features, namely $F_s$, from input set of images $X_i$.  We name it ``Blending'' because the input is a set of images $X_i$ and we concatenate them together along the channels. Thus, we end up with a $64*L*K$ input.
%It is in charge of combining the set of images $X_i$ to be represented as $64\times L\time K$ input tensor, so that the visual appearance feature $F_s$ can be produced.

%In our implementation, we have utilized the ResNet34~\cite{he2016deep} as the backbone of $S$ instead of VGG19~\cite{simonyan2014very} in GANwriting. We compare both of the convolutional architectures in Section~\ref{sec:exp}.

\subsection{Textual Content Encoder}

\begin{figure*}[ht!]
    \centering
    \includegraphics[width=0.85\linewidth]{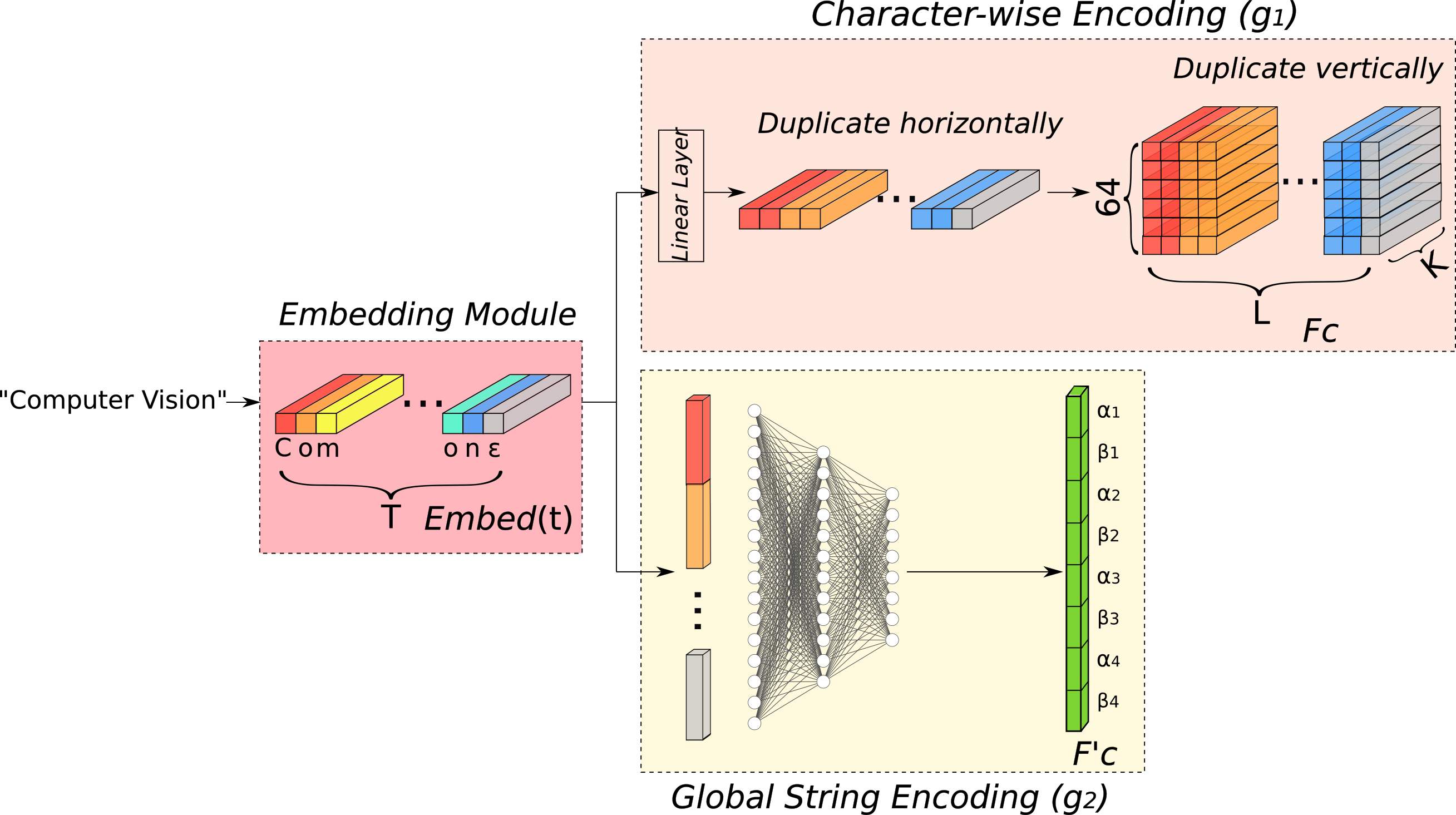}
    \caption{Architecture of the textual content encoder. It consists of an Embedding Module (red box), a Character-wise Encoding (orange box) and a Global String Encoding (yellow box). In the sequence of character embeddings, each vector is represented by a specific color.}
    \label{fig:cont_enc}
\end{figure*}

The textual content encoding process transits an input text string $t$ into textual content features $F_c$ and $F_c'$, as shown in Figure~\ref{fig:cont_enc}, and denoted as $ F_c, F_c' = C\left(t\right)$. Observe that the input text string $t$ is firstly padded with the empty symbol $\varepsilon$ to a fixed maximum string length $T$. Then, it goes through two pipelines: the character-wise embedding produces character-wise features $F_c$, and the global string encoding produces a global string description $F_c'$. These two pipelines have the textual content information that will be later combined with the visual appearance features during the generating process.

\subsubsection{Embedding Module}
As our method aims to generate any input sentence, including OOV words created from the predefined characters of alphabet $\mathcal{A}$, an embedding layer is applied on the input text string to extract character-level embedding features. Thus, each character $t_i \in t$ is mapped into a vector of size $n$ by means of the embedding function: 

\begin{align*}
    \operatorname{Embed} \colon \mathcal{A} &\to \mathbb{R}^n \\
    t_i &\mapsto \operatorname{Embed}(t_i),
\end{align*}

For the sake of simplicity and with abuse of notation, we will denote $\operatorname{Embed}(t)$ as the embedding of the whole string applied character by character.

\subsubsection{Character-wise Encoding}

%so as to produce a $T$-length embedding vectors, which is denoted as $Embed(t) \in \mathbb{R}^{T\cdot e}$ where $e$ is the embedding size. 

In order to properly combine the textual information with the style feature $F_s$ in the next step, the length $T$ should be aligned with the width of $F_s$. To achieve the alignment, each character embedding is repeated several times separately, and then all the chunks of repeated character embeddings are concatenated back together horizontally. To directly concatenate textual content feature $F_c$ and visual appearance feature $F_s$ in the next step, we also duplicate the horizontal character embeddings vertically. So the textual content feature $F_c$ ends up with the shape of ($64$, $L$, $K$) as shown in the upper part of Figure~\ref{fig:cont_enc} and denoted as $g_1$. Thus, we obtain the textual content feature $F_c$, which represents the local textual information and is obtained as $F_c = g_1(\operatorname{Embed}(t))$. Then both the content feature $F_c$ and the style feature $F_s$ are concatenated channel-wise.

\subsubsection{Global String Encoding}

Apart from injecting the character-wise information, an overall string information is helpful to guide the generating process as it gives a global coherence to the generation process. The character embedding $\operatorname{Embed}(t)$ is reshaped into a large one-dimensional vector of size $T \cdot n$. Then, a Multi-Layer Perceptron (MLP) $g_2$ is used to produce the global textual feature $F_{c}'$, as shown in the lower part of Figure~\ref{fig:cont_enc}. Thus, the global features are computed as $F_{c}' = g_2(\operatorname{Embed}(t))$, which is a one-dimensional vector. We first split it into 8 equally sized pieces, and then we use them as 4 pairs of $\alpha$ and $\beta$ parameters orderly, which will be used in AdaIN (refer to Equation~\ref{equ:adain}). 

\subsection{Generator}
The generator $G$ is in charge of combining the two sources of information: the visual appearance encoder and the textual content encoder. It consists of two residual blocks with AdaIN~\cite{huang2018multimodal} as the normalization layer, $4$ convolutional modules with nearest neighbor up-sampling and ReLU activations, and a final $\tanh(\cdot)$ activation layer. The global string information is equipped with the generator via AdaIN, which is formally defined as
\begin{equation}
\label{equ:adain}
	    \operatorname{AdaIN}\left(z, \alpha, \beta\right) = \alpha \left(\frac{z-\mu\left(z\right)}{\sigma\left(z\right)}\right) + \beta,
\end{equation}
where $z \in F$, $\mu$ and $\sigma$ are the channel-wise mean and standard deviations. The parameters $\alpha$ and $\beta$ are assigned with the splitting of $F_{c}^{'}$. Hence, the generative process is defined as
\begin{equation}
    \begin{aligned}
    \bar{x} &= H\left(t,X_i\right) = G\left( C\left(t\right), S\left(X_i\right)\right) = G\left( F_c, F_c', F_s\right)\\
    &= G\left( g_1\left(Embed(t)\right), g_2\left(Embed(t)\right), S\left(X_i\right)\right),
    \end{aligned}
\end{equation}
where $C$ is the textual content encoder and $S$ is the visual appearance encoder.

\begin{figure*}[t!]
    \centering
    \includegraphics[width=0.9\linewidth]{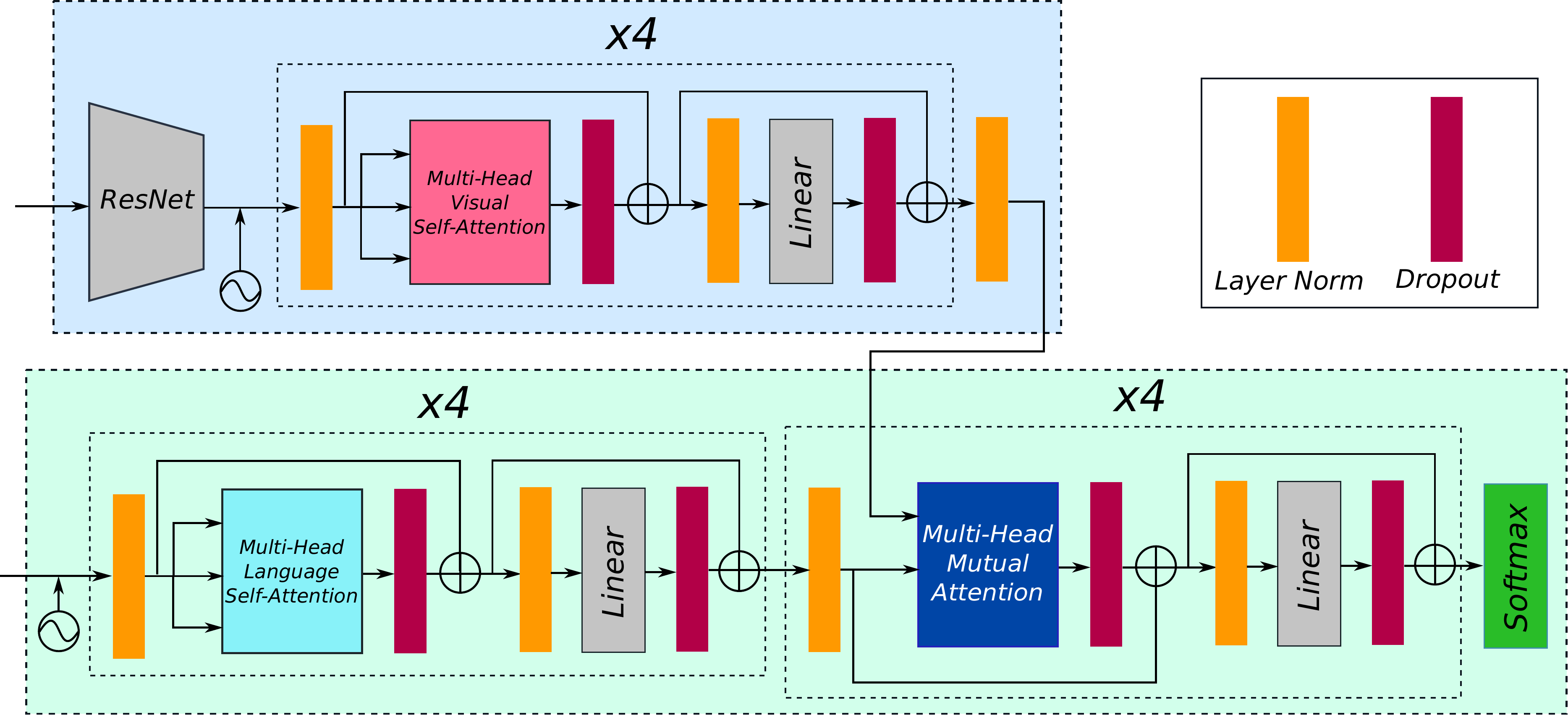}
    \caption{Architecture of the Transformer-based handwritten text recognizer. The upper part is the Encoder (blue color) and the lower part is the Decoder (green color).}
    \label{fig:transfo}
\end{figure*}

\subsection{Learning Objectives}
Three learning objectives are proposed to enforce different properties on the generated images $\bar{x} \in \bar{\mathcal{X}}$, where $\bar{\mathcal{X}}$ is the generated image space mimicking the visual appearance of real images $\mathcal{X}$. First, a discriminative loss $\mathcal{L}_d$ is in charge to ensure a realistic global appearance. Second, a writer classification loss $\mathcal{L}_w$ forces the generated samples to follow a specific appearance style. Finally a recognition loss $\mathcal{L}_r$ ensures the preservation of the textual content.

\subsubsection{Discriminative Loss}
Following the traditional GAN paradigm~\cite{goodfellow2014generative}, we propose a discriminative model $D$, which consists of one convolutional layer, six residual blocks, each of them with LeakyReLU activations and average poolings, and a final binary classification layer. Thus, given an input image, $D$ produces a binary output that is classified either as real ($1$) or fake ($0$). It does not take into consideration neither the visual appearance provided by $X_i$ nor the textual contents provided by $t$, and only focuses on the general visual appearance of the generated image $\bar{x}$ to look realistic. The discriminative loss $\mathcal{L}_d$ is formally defined as
\begin{equation}
    \mathcal{L}_d\left(H,D\right) = \mathbb{E}_{x \sim \mathcal{X}} \left[ \log\left(D\left(x\right)\right) \right] + \mathbb{E}_{\bar{x} \sim \bar{\mathcal{X}}}\left[\log\left(1-D\left(\bar{x}\right)\right)\right].
    \label{e:discriminator}
\end{equation}

\subsubsection{Visual Appearance Loss}
Assuming that each writer has an inherent writing style, we propose a writer classifier $W$ that follows the same architecture of $D$ by replacing the final binary classification layer with an $N$-classification layer, where $N$ is the number of writers in the training dataset. The writer classifier $W$ is optimized with real samples drawn from $\mathcal{X}$. This loss guides the generation of synthetic samples to align their styles with the given writer. Thus, the writer classifier acts as a style loss to provide diversity on the generated samples. The style loss $\mathcal{L}_w$ is formally defined as
\begin{equation}
    \mathcal{L}_w\left(H, W\right) = - \mathbb{E}_{x \sim \left\{ \mathcal{X},\bar{\mathcal{X}} \right\}} \left[ \sum_{i=1}^{\left|\mathcal{W}\right|} w_i \log\left(\hat{w}_i\right) \right],
    \label{e:style}
\end{equation}
where $\hat{w} = W(x)$ is the predicted probability distribution over writers in $\mathcal{W}$ and $w_i$ is the real writer distribution. Therefore, the generated samples should be classified as the same writer $w_i$ used to construct the input style conditioning image set $X_i$.

\subsubsection{Content Loss}
A handwritten text recognizer $R$ is used to ensure that the generated sample has the specific textual content, indicated by the input string $t$. Given that we generate text-lines, a robust recognizer for long sequences is needed. We adopt a Transformer-based recognizer~\cite{kang2020pay} that has recently shown good performance on full handwritten text-lines. 

The architecture of our Transformer-based HTR approach is shown in Figure~\ref{fig:transfo}. It also follows the encoder (upper part) and decoder (lower part) structure as proposed in~\cite{vaswani2017attention}. The encoder extracts high-level features from the input handwritten images, which consists of a ResNet and $4$ blocks of self-attention module and linear module with layer normalization and dropout. The decoder takes masked text strings as input~\cite{vaswani2017attention} so that the decoding only depends on predictions produced prior to the current character. In addition, the processing with characters is done in parallel, avoiding the recurrency of sequence-to-sequence models. Such a parallel processing of what used to be different time steps in sequence-to-sequence approaches drastically reduces the training time. The decoder consists of $4$ blocks of self-attention modules and $4$ blocks of mutual-attention modules, which provide an even more powerful ability to handle long sequence inputs than sequence-to-sequence approaches. The comparison between both of the sequence-to-sequence-based and Transformer-based methods is detailed in Section~\ref{sec:exp}.

The Kullback-Leibler divergence loss is used as the recognition loss at each time step. It is formally defined as:
\begin{equation}
    \mathcal{L}_r\left(H, R\right) = - \mathbb{E}_{x \sim \left\{ \mathcal{X},\bar{\mathcal{X}} \right\}} \left[ \sum_{i=0}^{L} \sum_{j=0}^{\left|\mathcal{A}\right|}  t_{i,j} \log\left(\frac{t_{i,j}}{\hat{t}_{i,j}}\right) \right],
    \label{e:content}
\end{equation}
where $\hat{t} = R(x)$; $\hat{t}_{i}$ being the $i$-th decoded character probability distribution by the recognizer, $\hat{t}_{i,j}$ being the probability of $j$-th symbol in $\mathcal{A}$ for $\hat{t}_{i}$, and $t_{i,j}$ being the real probability corresponding to $\hat{t}_{i,j}$. The empty symbol $\varepsilon$ is ignored in the loss computation.

\subsubsection{Joint Training Process}

The whole architecture is trained with the three proposed loss functions jointly in an end-to-end fashion as follows.
\begin{equation}
    \mathcal{L}(H, D, W, R) = \mathcal{L}_d(H, D) + \mathcal{L}_w(H, W) + \mathcal{L}_r(H, R),
    \label{e:final}
\end{equation}

\begin{equation}
    \min_{H,W,R} \max_D \mathcal{L}(H, D, W, R).
\end{equation}

The training strategy is further explained in Algorithm~\ref{alg:train}, where $\Theta = \{\Theta_{H}, \Theta_{D}, \Theta_{W}, \Theta_{R}\}$ represents the related network parameters and $\Gamma(\cdot)$ denotes the optimizer function. Even though the training process is end-to-end, the optimization process is performed in two steps. Firstly, we feed both real and generated samples together to the discriminator $D$, so that the discriminative loss can be obtained (line~\ref{alg:line:D}). Secondly, we only make use of real data to train the writer classifier $W$ and the text recognizer $R$, so that the visual appearance and content losses are obtained (line~\ref{alg:line:wr}). 
As $W$ and $R$ are optimized with only real data, they could be pre-trained independently as an initialization apart from the generative network $H$. However, the good performance of $W$ and $R$ may unbalance the three losses in the early training steps, which could make the whole network hard to train. Thus, we initialize all the network parameters from scratch and jointly train them altogether. The network parameters $\Theta_D$ are optimized by gradient ascent following the GAN paradigm whereas the parameters $\Theta_W$ and $\Theta_R$ are optimized by gradient descent. Finally, the overall generator loss is computed following Equation~\ref{e:final} where only the generator parameters $\Theta_H$ are optimized (line~\ref{alg:line:H}).

\begin{algorithm}[t!]
\hspace*{\algorithmicindent} \textbf{Input:} \small{Input data $\{\mathcal{X}, \mathcal{Y}, \mathcal{W}\}$; alphabet $\mathcal{A}$; max training iterations $Itr$} \\ 
\hspace*{\algorithmicindent} \textbf{Output: } \small{Networks parameters $\Theta = \{\Theta_{H}, \Theta_{D}, \Theta_{W}, \Theta_{R}\}$.}
\begin{algorithmic}[1]
\Repeat
\State Get style and content mini-batches $\{X_i, w_i\}_{i=1}^{N_{Batch}}$ and $\{t^i\}_{i=1}^{N_{Batch}}$
\State $\mathcal{L}_d \leftarrow $ Eq.~4 \label{alg:line:D} \algorithmiccomment{Real and generated samples $x \sim \{\mathcal{X}, \bar{\mathcal{X}}\}$}
\State $\mathcal{L}_{w,r} \leftarrow $ Eq.~5 + Eq.~6 \label{alg:line:wr} \algorithmiccomment{Real samples $x \sim \mathcal{X}$}
\State $\Theta_D \leftarrow \Theta_D + \Gamma(\nabla_{\Theta_D}\mathcal{L}_d)$
\State $\Theta_{W,R} \leftarrow \Theta_{W,R} - \Gamma(\nabla_{\Theta_{W,R}}\mathcal{L}_{w,d})$
\State $\mathcal{L} \leftarrow $ Eq.~\ref{e:final} \algorithmiccomment{Generated samples $x \sim \bar{\mathcal{X}}$}
\State $\Theta_H \leftarrow \Theta_H - \Gamma(\nabla_{\Theta_{H}}\mathcal{L})$ \label{alg:line:H}
\Until{Max training iterations $Itr$}
\end{algorithmic}
\caption{Training algorithm for the proposed model.} \label{alg:train}
\end{algorithm}
%$t_i$ denotes the $i$-th character on the input text $t$.

% \begin{enumerate}
%     \item Number of characters less than $23$ and image length less than $600$ pixels.
%     \item Number of characters less than $48$ and image length less than $1200$ pixels.
%     \item Number of characters less than $88$ and image length less than $2160$ pixels.
% \end{enumerate}

\section{Variable-length Fr\'{e}chet Inception Distance (vFID)}
\label{sec:vfid}

\begin{figure}[h]
    \centering
    %\resizebox{\linewidth}{!}{
    %\hspace*{-2.5cm}
    \begin{tabular}{cc}
    \includegraphics[width=0.46\linewidth]{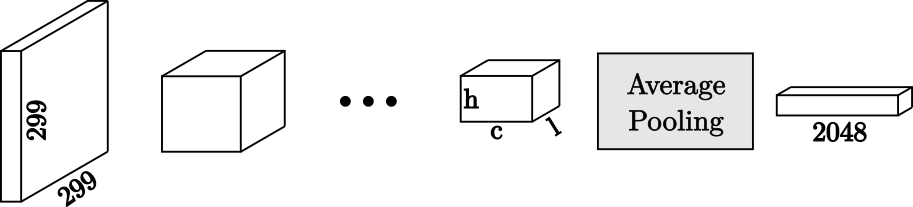} &
    \includegraphics[width=0.46\linewidth]{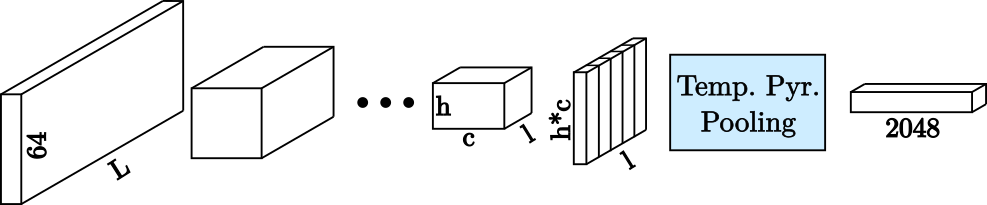}\\
     (a) & (b)\\
\end{tabular}
%\hspace*{-2.5cm}
%}%
    \caption{(a) Inception module of FID with Average Pooling, (b) Updated Inception module of vFID with Temporal Pyramid Pooling.}
    \label{fig:tpp}
\end{figure}

In order to evaluate the quality of the generated images by GANs, there is a commonly used metric, namely Fr\'{e}chet Inception Distance (FID)~\cite{dowson1982frechet}. FID is used to evaluate the similarity between the generated images and the real ones. This is achieved by calculating the distance between two feature vectors, which are obtained from two image sets, the generated images and the real ones, respectively. FID has been widely used in evaluating the performance of GANs at image generation. This metric follows two steps: first, it extracts features from an InceptionV3 network~\cite{heusel2017gans} while keeping activations of the last pooling layer, which is pretrained on the ImageNet dataset~\cite{deng2009imagenet}; then, it calculates the distance between the feature vectors. Even though FID has been widely used for the evaluation of the generated natural scene images, it is not well suited for handwritten image data. The main drawbacks in such case are (i) the ImageNet dataset consists of natural scene image samples that have very few common features with handwritten text images; (ii) the InceptionV3 model used by the FID requires a fixed size input, which could not handle the variable-length scenario of handwritten text images. 
Thus, we introduce a novel version of FID, namely vFID (Variable-length Fr\'{e}chet Inception Distance), specially suited for such variable-length images such as handwritten text images. Similarly to the original FID, the proposed metric vFID share the same InceptionV3 network as the convolutional backbone. However, instead of the average pooling used by the FID, we first reshape the convolutional feature into a $2$-dimensional feature map which is then fed into a Temporal Pyramid Pooling (TPP) layer~\cite{wang2016temporal} as shown in Figure~\ref{fig:tpp}. TPP is especially useful when the input is a variable-length sequence of features, which is the case for handwritten text-line images. Based on the pretrained InceptionV3, we fine-tune the vFID model with the IAM dataset by fitting a writer classifier. When applying the vFID metric, the input images should be resized to have $64$ pixels height while preserving the aspect ratio. Thus, the variable resulting width is denoted by $L$. We calculate the vFID values for each input image without adding paddings. Thus, vFID is not affected by batching and different image widths.

\begin{figure}[h]
    \centering
    \begin{tabular}{cc}
        \includegraphics[width=0.48\linewidth]{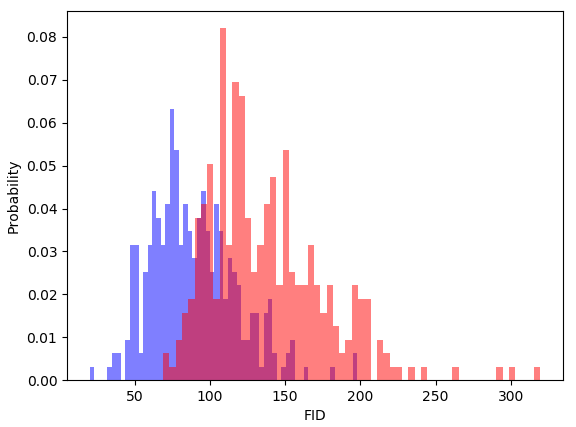} 
        &\includegraphics[width=0.48\linewidth]{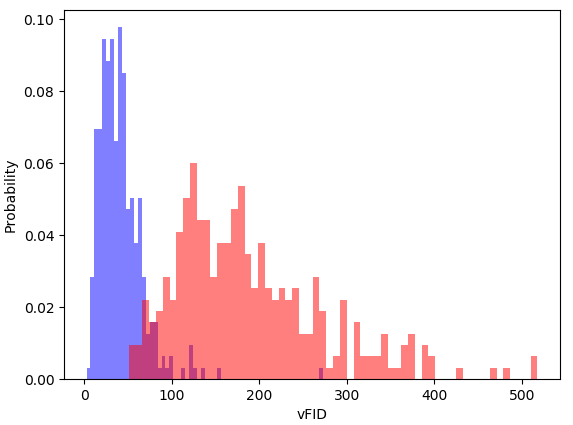}  \\
        (a) & (b)\\
    \end{tabular}
    \caption{Histogram of FID (a) and vFID (b). The x-axis indicates the FID/vFID values, and the y-axis indicates the counts. The FID/vFID between subsets of samples in the same writer is shown in blue, and between different writers in red. The distribution of blue and red should be apart as far as possible. Both histograms are normalized to sum up to one.}
    \label{fig:histo}
\end{figure}

To achieve a valid metric performance and a fair comparison for both vFID and FID, we first reuse the Inception V3 network that is pretrained on ImageNet dataset, and then, we fine-tune both vFID and FID models on the IAM datasets towards a writer classification problem. Once they are properly trained, we can then evaluate the generated image quality through the metrics.
The performance comparison of FID and vFID for the IAM dataset is shown in Figure~\ref{fig:histo}. The blue distribution indicates the performance for the same writer, while the red one indicates the performance for a different writer pair. The lower value of the FID/vFID, the better similarity is obtained. Therefore, we aim to achieve a robust metric that produces a lower value for the same writer (blue) and a higher value for the different writers (red). In Figure~\ref{fig:histo}(a), we observe that the performance of FID do not have a good behaviour since it has a big overlapping area, so that it cannot provide a reasonable judgement on the performance of the generated text. Contrary, our proposed vFID in Figure~\ref{fig:histo}(b) could provide a more trustful measure. 

%The Blue distribution is done by randomly splitting $X_i$ into half-half and then applying FID/vFID, while the red one is done by randomly selecting $2$ different writers $w_a$ and $w_b$ and then applying FID/vFID on the $2$ sets $X_a$ and $X_b$. The behaviour of a good metric should produce a low value for the samples of same writer and a higher value for that of different writers. 

%%%%%%%%%%%%%%%%%%%%%%%%%%%%%%%%%%%%%%%%%%%%%%%%%%%%%%%%%%%%%%%%%%%%%%
%%%%%%%%%%%%%%%%%%%%%%%%%%%%%%%%%%%%%%%%%%%%%%%%%%%%%%%%%%%%%%%%%%%%%%
\section{Experiments}
\label{sec:exp}
In this section, we present the extensive evaluation of our proposed approach. First, we perform several ablation studies on the key modules to find the best balance between performance and efficiency. Then, we demonstrate qualitative and quantitative results on synthetically generated images. Finally, we make use of the generated samples to boost the HTR performance in different experimental settings.  

\subsection{Datasets and Metrics}
\label{sec:dataset}
\begin{figure}[t!]
    \centering
    \begin{tabular}{c}
        \includegraphics[height=0.8cm]{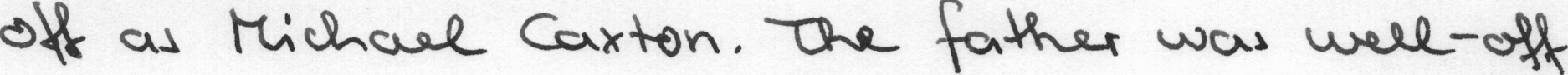}\\
        \includegraphics[height=0.8cm]{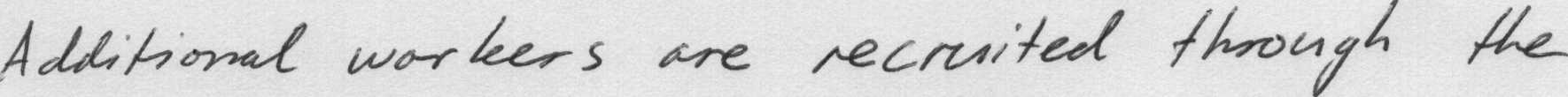}\\ 
        (a)\\
        \includegraphics[height=0.95cm]{images/example/train2011-0-5.png}\\
        \includegraphics[height=0.95cm]{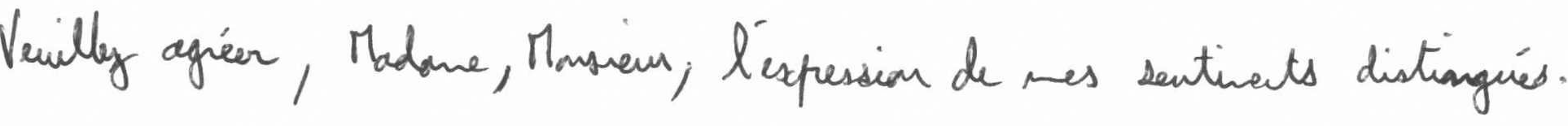}\\
        (b)\\
        \includegraphics[height=0.78cm]{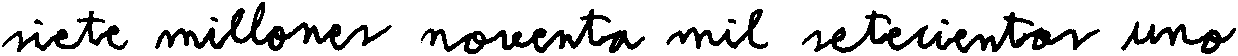}\\
        \includegraphics[height=0.78cm]{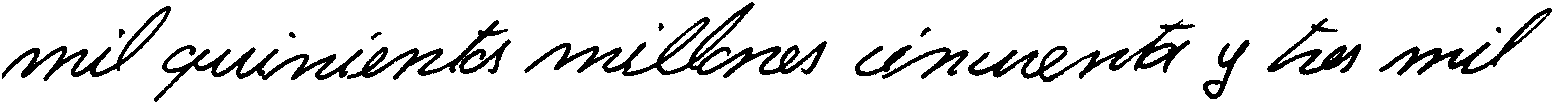}\\
        (c)\\
    \end{tabular}
    \caption{Examples of the IAM, Rimes and Spanish Numbers datasets are shown in (a), (b) and (c), respectively.}
    \label{fig:example}
\end{figure}

The IAM offline dataset~\cite{marti2002iam}, the Rimes dataset~\cite{augustin2006rimes} and the Spanish Numbers dataset~\cite{toselli2004integrated} are utilized in our experiments as shown in Table~\ref{tab:data}. However, our proposed generative method is only trained with IAM dataset. All the three datasets are utilized in the HTR experiments. Examples of the three datasets are shown in Figure~\ref{fig:example}.

\begin{table}[h!]
    \caption{Overview of the datasets used in our HTR experiments: Number of text-lines used for training, validation and test sets, and number of writers.}
    \label{tab:data}
    \centering
    \begin{tabular}{llllll}
    \toprule
    Dataset & Train & Val. & Test & Writers & Language\\
    \midrule
    IAM~\cite{marti2002iam} & $6482$ & $976$ & $2914$ & $657$ & English\\
    Rimes~\cite{augustin2006rimes} & $11333$ & $-$ & $778$ & $1300$ & French\\
    Spanish Num.~\cite{toselli2004integrated} & $298$ & $-$ & $187$ & $30$ & Spanish\\
    \bottomrule
    \end{tabular}
\end{table}

WikiText-103~\cite{merity2016pointer} is chosen to be our external text corpus when selecting random text strings as textual input. As in the case of images, we select texts in WikiText-103 from one word to $N_t$ words to create sentences. We end up with 3.6 million text-lines with number of characters varying from 1 to 88. 

The \emph{Character Error Rate} (CER) and the \emph{Word Error Rate} (WER)~\cite{frinken2014continuous} are the performance measures. The CER is computed as the Levenshtein distance, which is the sum of the character substitutions ($S_c$), insertions ($I_c$) and deletions ($D_c$) that are needed to transform one string into the other, divided by the total number of characters in the groundtruth ($N_c$). Formally, 
    \begin{equation}
	   CER = \frac{S_c + I_c + D_c}{N_c}
	\end{equation}
Similarly, the WER is computed as the sum of the word substitutions ($S_w$), insertions ($I_w$) and deletions ($D_w$) that are required to transform one string into the other, divided by the total number of words in the groundtruth ($N_w$). Formally,
	\begin{equation}
	   WER = \frac{S_w + I_w + D_w}{N_w}
	\end{equation}
	
\subsection{Curriculum Learning Strategy}
The IAM dataset is used to train our generative method. It is a multi-writer dataset in English, which consists of $1,539$ scanned pages written by 657 writers, as detailed in Table~\ref{tab:data}. Since we can access to the groundtruth of the training data, including the bounding-boxes at word level, we could enlarge the training set using the N-gram cropping strategy. For example, given a sequence of words, we can iteratively crop out one-word, two-words, and so on until $N$-word sub-lines, where $N$ is the maximum number of words in the given text-line. Thus, given the normalized height of $64$ pixels, we end up with $598,489$ images with variable lengths from $64$ to $2160$ pixels and the number of characters from $1$ to $88$, where $2160$ and $88$ are the maximum image length and text length for IAM dataset, respectively. 

To achieve a better handwriting generation with fine-grained details, we make use of curriculum learning strategy by splitting training data into $3$ categories as shown in Table~\ref{tab:category}, from shorter to longer sentences. We start the training with data of Category $1$ from scratch, then we fine-tune with data of Category $2$, and finally we fine-tune with data of Category $3$. The training is done step by step with increasing difficulty in the sense of image and text length. Note that the data used in the previous training step does not appear in the next fine-tuning step considering the training speed. In practice, the second and third steps just need to be fine-tuned for few epochs.

\begin{table}[t!]
    \caption{Three categories of the IAM offline dataset, from short to long text-lines.}
    \label{tab:category}
    \centering
    \begin{tabular}{ccc}
    \toprule
    Category & Num. of chars. & Image length\\
    \midrule
    $1$ & $1-24$ & $64-600$\\
    $2$ & $24-48$ & $600-1200$\\
    $3$ & $48-88$ & $1200-2160$\\
    \bottomrule
    \end{tabular}
\end{table}

\subsection{Implementation Details}
The experiments were run using PyTorch~\cite{paszke2017automatic} on a single NVIDIA RTX6000 GPU. As there are three objective modules, we set the learning rate of the discriminator and the generator to be $1\cdot 10^{-4}$, whereas the ones of writer classifier and recognizer are $1\cdot 10^{-5}$. The training process is optimized with Adam optimizer and a batch size of $4$. 

\subsection{Ablation Study}

% \begin{itemize}
%     \item \sout{w/ o w/o Style Filling Module, also visualize styled features}
%     \item \sout{Style encoder: VGG-19 or ResNet-34, performance and speed}
%     \item character wise encoding (appearance of each char), global string encoding (natural cursive connections between chars). 1-0, 0-1, 1-1
%     \item recognizer: seq2seq vs transformer, increasing the number of chars
% \end{itemize}

As explained in Section~\ref{sec:vfid}, the vFID can measure the similarity of both real and generated images in a better way than the original FID, so we make use of vFID to choose the convolutional architecture and the recognizer.

\begin{figure*}[t!]
    \centering
    \hspace*{-2.5cm}
    \begin{tabular}{ll}
    \toprule
    Style (w/o) & \includegraphics[height=0.45cm]{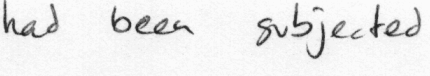}\\
    \textbf{Output} & \includegraphics[height=0.45cm]{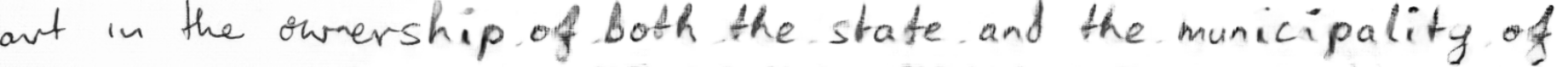}\\
    \midrule
    Style (w/) & \includegraphics[height=0.45cm]{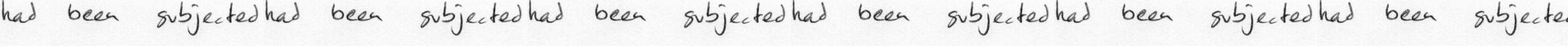}\\
    \textbf{Output} & \includegraphics[height=0.45cm]{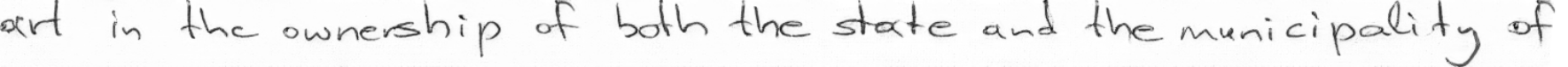}\\
    \bottomrule
    \end{tabular}
    \hspace*{-2.5cm}
    \caption{Comparison of the generated results for the same text string ``art in the ownership of both the state and the municipality of'' without (upper) and with (lower) the periodic padding process.}
    \label{fig:sty_fill}
\end{figure*}

First, we compare the generated samples with and without Periodic Padding Module as shown in Figure~\ref{fig:sty_fill}. The style input is randomly selected from a specific writer, and it may be a shorter image than what we expect to generate. In the upper part of Figure~\ref{fig:sty_fill}, the style input is padded with $0$ to the maximum width, so that the generated image suffers from the style collapse problem in the corresponding padded area. Contrary, in the lower part of Figure~\ref{fig:sty_fill}, with the periodic padding process, the style image has been extended to the maximum width that is sure of covering all the possibly generated area. Thus, the generated sample keeps the consistency in the visual appearance from the first character until the end.

Second, we modify the convolutional layers from VGG19 to ResNet34, and study the performance and training speed. The performance is evaluated on the generated samples based on the style information from IAM test set and content information from a subset of WikiText-103. The models are trained until $500$ epochs. Speed is the total time for a forward and backward pass. From Table~\ref{tab:conv}, we observe that ResNet34 achieves a higher training speed while obtaining a slightly better performance.

\begin{table}[h!]
    \caption{Ablation study for Convolutional layers on the IAM test set.}
    \label{tab:conv}
    \centering
    \begin{tabular}{lcc}
    \toprule
    Conv. & vFID & Speed (ms)\\
    \midrule
    VGG19 & 115.46 & 144.13\\
    ResNet34 & \textbf{114.39} & \textbf{136.80}\\
    \bottomrule
    \end{tabular}
\end{table}

\begin{figure*}[ht!]
    \centering
    \resizebox{\linewidth}{!}{
    %\hspace*{-1cm}
    \begin{tabular}{lc}
    \toprule
    Style A: & \includegraphics[height=0.5cm,valign=c]{images/interp/l01-023-05.png}\\
    \midrule
    0.0: & \includegraphics[height=0.5cm,valign=c]{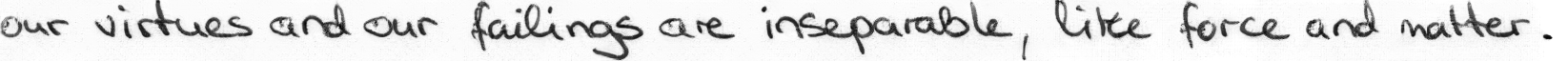}\\
    0.1: & \includegraphics[height=0.5cm,valign=c]{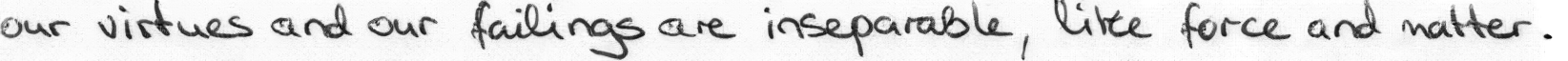}\\
    0.2: & \includegraphics[height=0.5cm,valign=c]{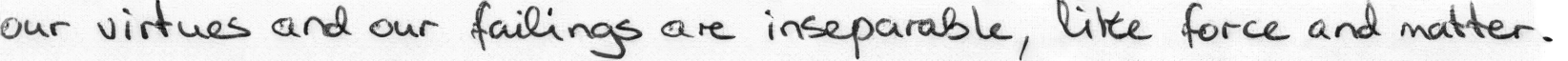}\\
    0.3: & \includegraphics[height=0.5cm,valign=c]{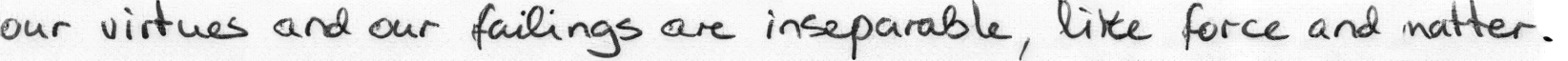}\\
    0.4: & \includegraphics[height=0.5cm,valign=c]{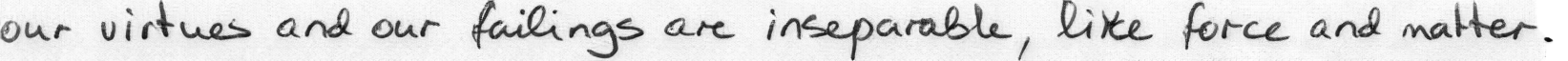}\\
    0.5: & \includegraphics[height=0.5cm,valign=c]{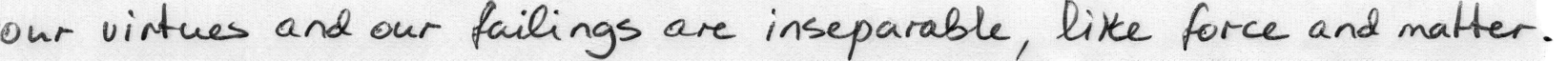}\\
    0.6: & \includegraphics[height=0.5cm,valign=c]{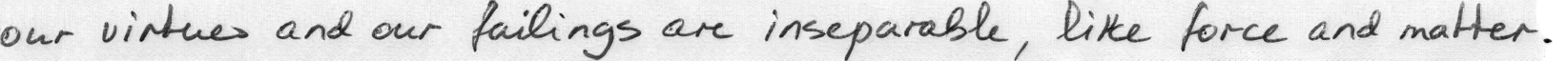}\\
    0.7: & \includegraphics[height=0.5cm,valign=c]{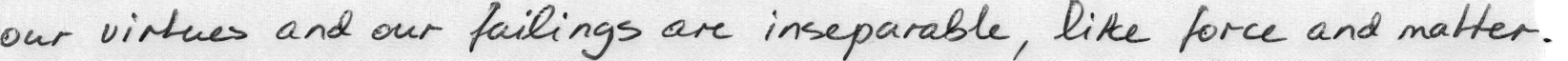}\\
    0.8: & \includegraphics[height=0.5cm,valign=c]{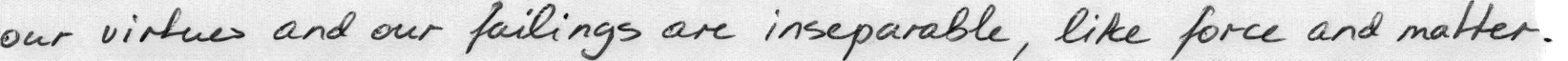}\\
    0.9: & \includegraphics[height=0.5cm,valign=c]{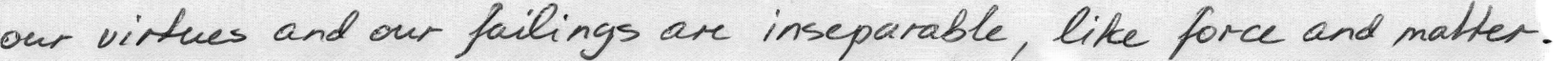}\\
    1.0: & \includegraphics[height=0.5cm,valign=c]{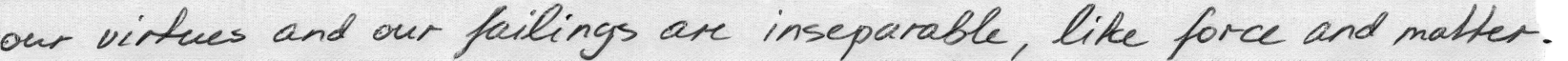}\\
    \midrule
    Style B: & \includegraphics[height=0.5cm,valign=c]{images/interp/h07-080-04.png}\\
    \bottomrule
\end{tabular}
%\hspace*{-1cm}
}
    \caption{Example of interpolations in the style latent space.}
    \label{fig:interpA}
\end{figure*}

% Third, we study on the content information injection process. As detailed in Section~\ref{sec:method}, two pipelines are utilized to equip the textual content feature into the generating process: concatenation of character-wise embeddings and AdaIn of global string encoding. 

% \begin{table}[t!]
%     \caption{Effect of partial and global content information injection.}
%     \label{tab:con_inject}
%     \centering
%     \begin{tabular}{cccc}
%         \toprule
%         Char. Embed. & Global Enc. & vFID & Sample\\
%         \midrule
%         \checkmark & $-$ & \\
%         $-$ & \checkmark & \\
%         \checkmark & \checkmark & \\
%         \bottomrule
%     \end{tabular}
% \end{table}

\begin{table}[h!]
    \caption{vFID performance on generating different length of images for the sequence-to-sequence and Transformer-based HTR methods. The lower the value, the better the performance.}
    \label{tab:rec}
    \centering
    \begin{tabular}{ccc}
        \toprule
        \multirow{2}{*}{Method} & \multicolumn{2}{c}{\textbf{Num. of chars to be generated}}\\
         & 1-10 (words) & 1-90 (lines)\\
        \midrule
        Seq2Seq & \textbf{136.51} &  249.94\\
        Transformer & 146.74 & \textbf{114.39}\\
        \bottomrule
    \end{tabular}
\end{table}

Third, we analyze the effect of replacing the sequence-to-sequence recognizer with the Transformer-based recognizer. Based on the number of characters to be rendered in the generated samples, we have two categories: words with 1 up to 10 characters and text-lines with 1 up to 90 characters, as shown in Table~\ref{tab:rec}. From the Table we observe that sequence-to-sequence-based method performs well at word level but it significantly degrades when extending to text-lines. Contrary, the Transformer-based HTR method achieves a better performance when dealing with longer text sequences. Since the transformer network has the ability of dealing with long-term dependencies, it becomes more powerful to control the textual content of the generated samples. 

Finally, we analyze the two schemes, either character-wise encoding ($g_1$) or global string encoding ($g_2$), to merge with the visual appearance feature as condition for the generation process. As shown in Table~\ref{tab:local_global}, the best performance is achieved with the use of both local and global encodings. Thus, the two schemes are utilized altogether in our model.

\begin{table}[h!]
    \caption{Ablation study on the use of character-wise encoding (local feature) and global string encoding (global feature). vFID values are calculated on the IAM at word-level (1-10 characters).}
    \label{tab:local_global}
    \centering
    \begin{tabular}{ccc}
    \toprule
    Local ($g_1$) & Global ($g_2$) & vFID\\
    \midrule
    \checkmark & & 162.38\\
     & \checkmark & 149.07\\
    \checkmark & \checkmark & \textbf{146.74}\\
    \bottomrule
    \end{tabular}
\end{table}

\subsection{Latent Space Interpolation}
\begin{figure*}[t!]
    \centering
    \resizebox{\linewidth}{!}{
    %\hspace*{-1cm}
    \begin{tabular}{lc}
    \toprule
    Style C: & \includegraphics[height=0.5cm,valign=c]{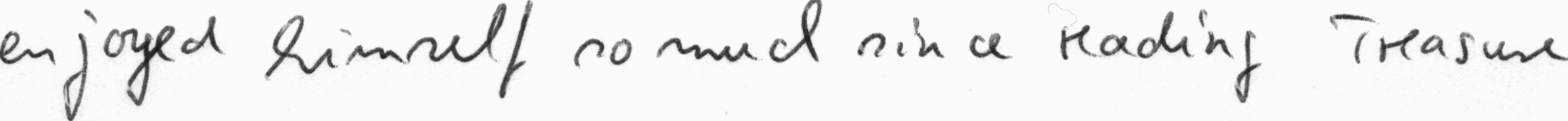}\\
    \midrule
    0.0: & \includegraphics[height=0.5cm,valign=c]{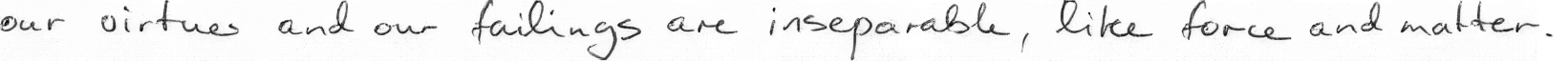}\\
    0.1: & \includegraphics[height=0.5cm,valign=c]{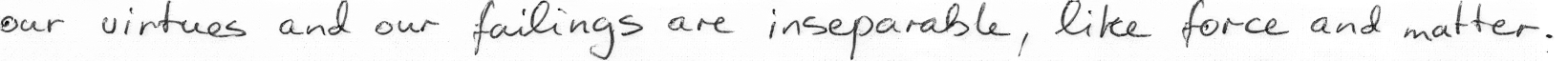}\\
    0.2: & \includegraphics[height=0.5cm,valign=c]{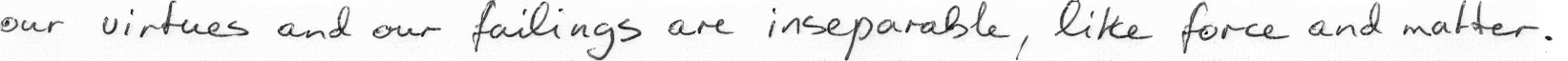}\\
    0.3: & \includegraphics[height=0.5cm,valign=c]{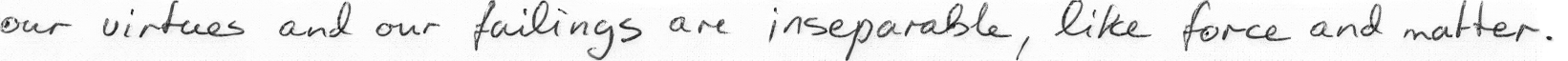}\\
    0.4: & \includegraphics[height=0.5cm,valign=c]{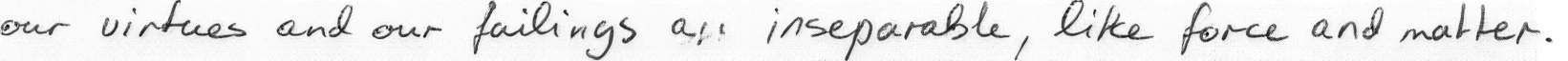}\\
    0.5: & \includegraphics[height=0.5cm,valign=c]{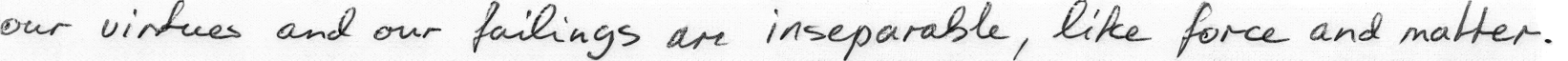}\\
    0.6: & \includegraphics[height=0.5cm,valign=c]{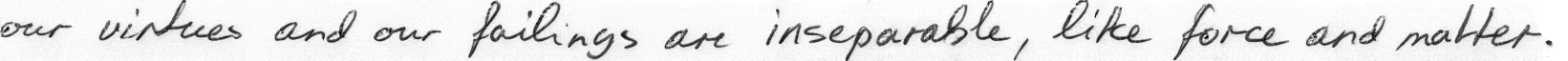}\\
    0.7: & \includegraphics[height=0.5cm,valign=c]{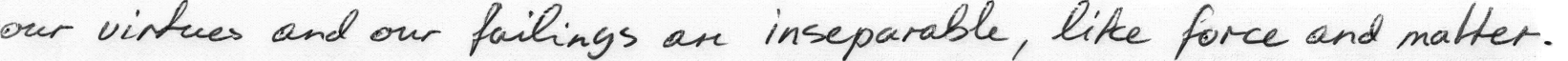}\\
    0.8: & \includegraphics[height=0.5cm,valign=c]{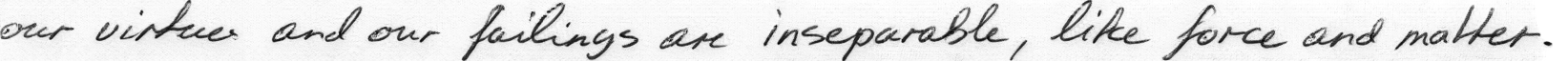}\\
    0.9: & \includegraphics[height=0.5cm,valign=c]{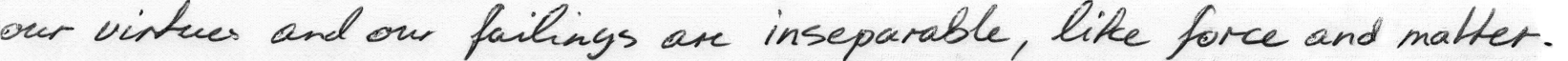}\\
    1.0: & \includegraphics[height=0.5cm,valign=c]{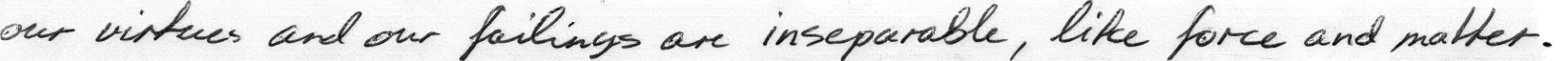}\\
    \midrule
    Style D: & \includegraphics[height=0.5cm,valign=c]{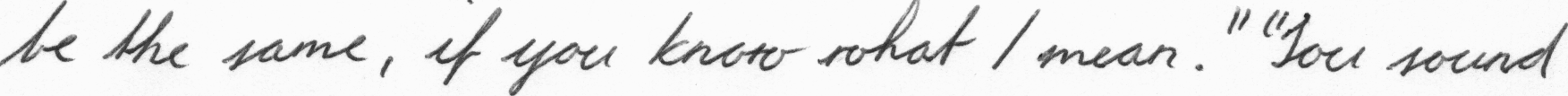}\\
    \bottomrule
\end{tabular}
%\hspace*{-1cm}
}
    \caption{Example of interpolations in the style latent space.}
    \label{fig:interpB}
\end{figure*}

Once the system is trained, the generator $G$ has learned a map in the handwriting style latent space. Each writer corresponds to a point in this latent space and different writers are connected in a continuous way. Thus, we can explore it by randomly choosing two writers and try to traverse between the corresponded two points in the style latent space as shown in Figures~\ref{fig:interpA} and~\ref{fig:interpB}. The first and last rows show the real samples from writer A and B, respectively. The samples in between are synthetically generated ones that try to traverse from writer A to B. The rendered text is the quote of ``our virtues and our failings are inseparable, like force and matter'' from Nikola Tesla, which has not been seen during training. 

\subsection{Handwritten Text-line Generation}

\begin{figure*}[t!]
    \centering
    %\resizebox{\linewidth}{!}{
    \hspace*{-2.5cm}
    \begin{tabular}{lc}
    \toprule
    Style Input: & \includegraphics[height=0.55cm,valign=c]{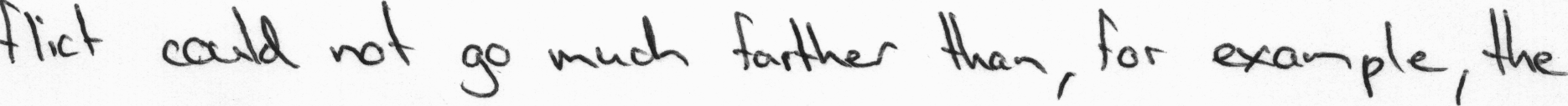}\\
    \midrule
    Text En: & ``the progressive development of man is vitally dependent on invention.''\\
    \textbf{Output:} & \includegraphics[height=0.45cm,valign=c]{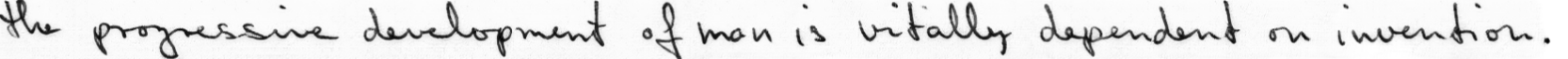}\\
    \midrule
    Text De: & ``die fortschreitende entwicklung des Menschen hangt entscheidend von der erfindung ab.''\\
    \textbf{Output:} & \includegraphics[height=0.45cm,valign=c]{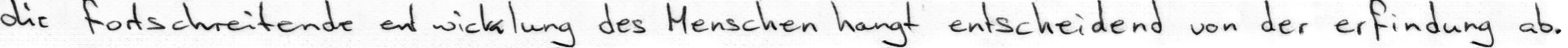}\\
    \midrule
    Text Fr: & ``le developpement progressif de l'homme depend de facon vitale de l'invention.''\\
    \textbf{Output:} & \includegraphics[height=0.45cm,valign=c]{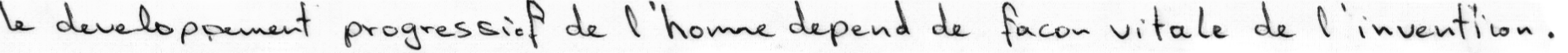}\\
    \midrule
    Text Es: & ``el desarrollo progresivo del hombre depende vitalmente de la invencion.''\\
    \textbf{Output:} & \includegraphics[height=0.45cm,valign=c]{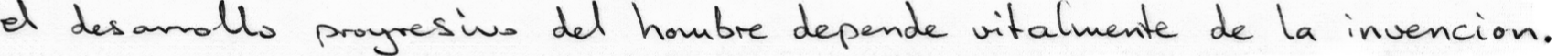}\\
    \bottomrule
\end{tabular}
\hspace*{-2.5cm}
%}%
    \caption{Generation on varied multi-lingual texts.}
    \label{fig:generation_sty}
\end{figure*}

\begin{figure*}[t!]
    \centering
    %\resizebox{\linewidth}{!}{
    \hspace*{-2cm}
    \begin{tabular}{lc}
    \toprule
    Text Input: & ``the progressive development of man is vitally dependent on invention.''\\
    \midrule
    Style A: & \includegraphics[height=0.55cm,valign=c]{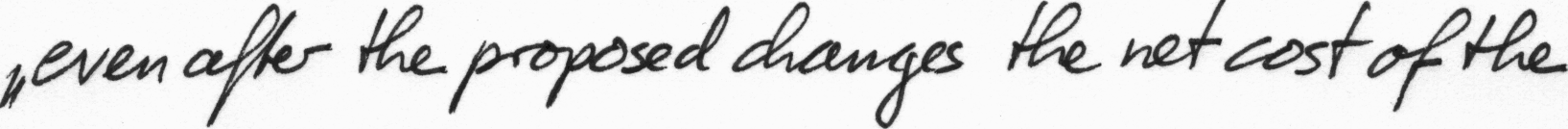}\\
    \textbf{Output:} & \includegraphics[height=0.45cm,valign=c]{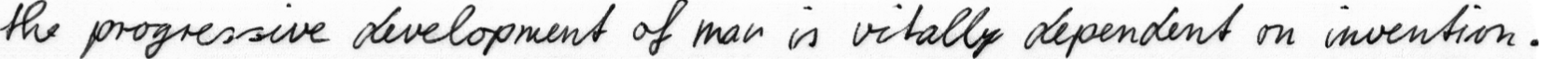}\\
    \midrule
    Style B: & \includegraphics[height=0.5cm,valign=c]{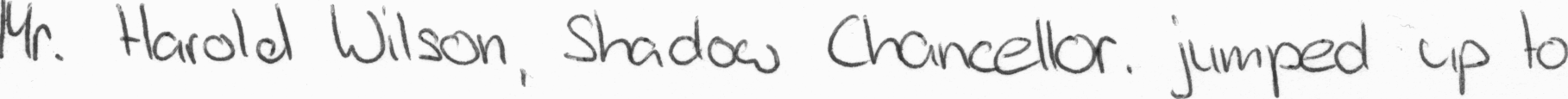}\\
    \textbf{Output:} & \includegraphics[height=0.5cm,valign=c]{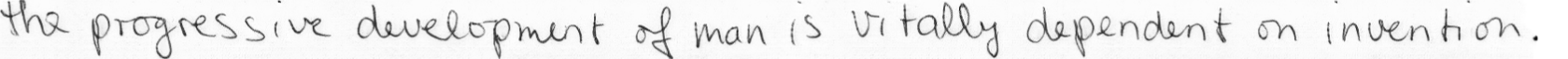}\\
    \midrule
    Style C: & \includegraphics[height=0.53cm,valign=c]{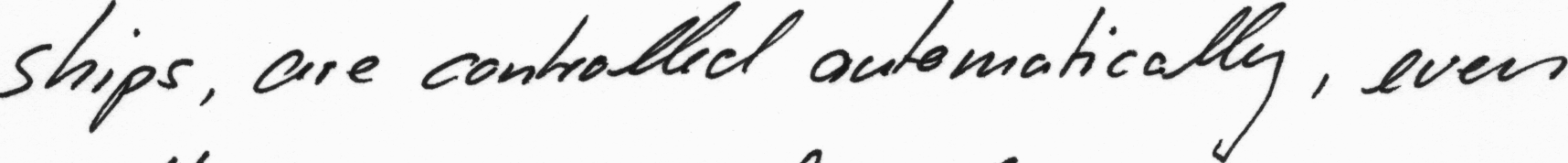}\\
    \textbf{Output:} & \includegraphics[height=0.47cm,valign=c]{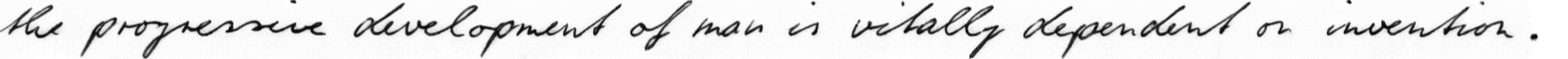}\\
    \midrule
    Style D: & \includegraphics[height=0.47cm,valign=c]{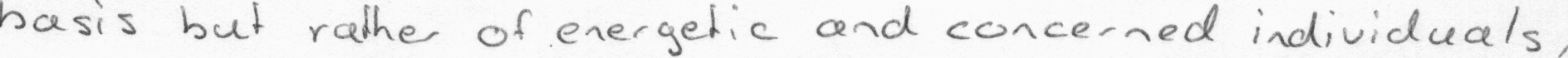}\\
    \textbf{Output:} & \includegraphics[height=0.52cm,valign=c]{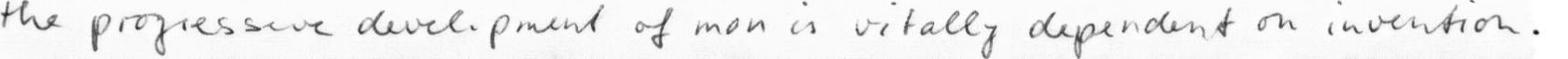}\\
    \bottomrule
\end{tabular}
\hspace*{-2cm}
%}
    \caption{Generation of varied styles.}
    \label{fig:generation_txt}
\end{figure*}

For the qualitative experiments, we show the results in two cases. First, given a same writing style, we try to generate samples with different text strings. Second, given a specific text string, we try to generate samples in different writing styles. The first case is shown in Figure~\ref{fig:generation_sty}. The text string is the quote of ``the progressive development of man is vitally dependent on invention.'' from Nikola Tesla. We translate it into German, French and Spanish while replacing special characters with the corresponding letters (e.g. ``\'e'' to ``e''). In Figure~\ref{fig:generation_sty}, the first row is a sample of the style input, and the following rows are (text string, synthetically generated sample) pairs. By showing the generation on different languages, our method proves to be not restricted to a training corpus nor a language model. Thus, this method can be applied to generate any OOV words and sentences. The second case is shown in Figure~\ref{fig:generation_txt}. The first row is the text string input, and the following rows are (handwriting style sample, synthetically generated sample) pairs. From the results we observe that our method has the ability to generate text-line samples with diverse writing styles.

Furthermore, we show a comparative with the state-of-the-art methods on handwritten text generation in Table~\ref{tab:alonso}. In our previous work~\cite{kang2020ganwriting}, we have conducted a human evaluation study to show that the generated samples are indistinguishable by humans. However, in this paper we focus on the improvement of HTR performance, so the interested reader is referred to our previous publication for details on the human evaluation study.

\newlength{\myheight}
\setlength{\myheight}{2.5ex}
\begin{table*}[ht!]
\caption{Qualitative comparison with Alonso~\etal~\cite{alonso2019adversarial}, Fogel~\etal~\cite{fogel2020scrabblegan}, and Davis~\etal~\cite{davis2020text}. These images are cropped from their papers. Three random writing styles are selected in our results, where Style $A$ is IAM writer $583$, Style $B$ is IAM writer $281$, and Style $C$ is Rimes writer $lot\_13\_01258$.}
    \label{tab:alonso}
    \centering
    %\begin{tabular}{c@{\hskip 8pt}c cccccc c c@{\hskip 8pt} c}
    \hspace*{-1cm}
    \begin{tabular}{ccccccccc}
    \toprule
     \multirow{3}{*}{Content} && \multirow{3}{*}{\cite{alonso2019adversarial}} & \multirow{3}{*}{\cite{fogel2020scrabblegan}} & \multirow{3}{*}{\cite{davis2020text}} && \multicolumn{3}{c}{Ours}\\
     \cmidrule{7-9}
     &&&&&& Style $A$ & Style $B$ & Style $C$\\
     &&&&&&  \includegraphics[height=\myheight,valign=c]{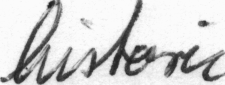} &  \includegraphics[height=\myheight,valign=c]{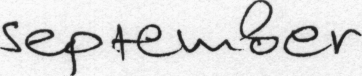} &  \includegraphics[height=2.2ex,valign=c]{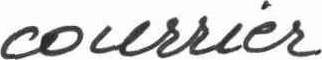}\\
     \cmidrule[\lightrulewidth]{1-1} \cmidrule[\lightrulewidth]{3-5} \cmidrule[\lightrulewidth]{7-9}
        \texttt{"olibus"}&&
        \includegraphics[height=\myheight,valign=c]{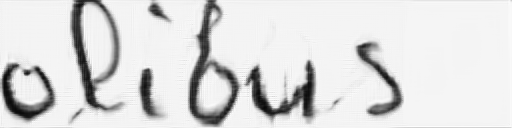} &
        \includegraphics[height=\myheight,valign=c]{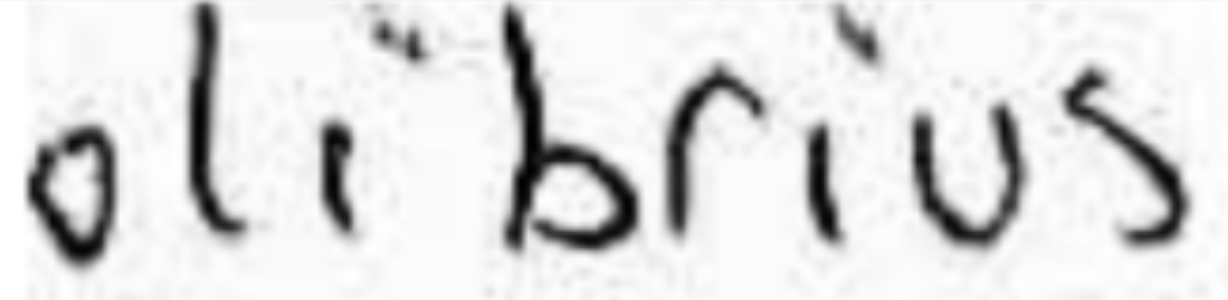} &
        \includegraphics[height=\myheight,valign=c]{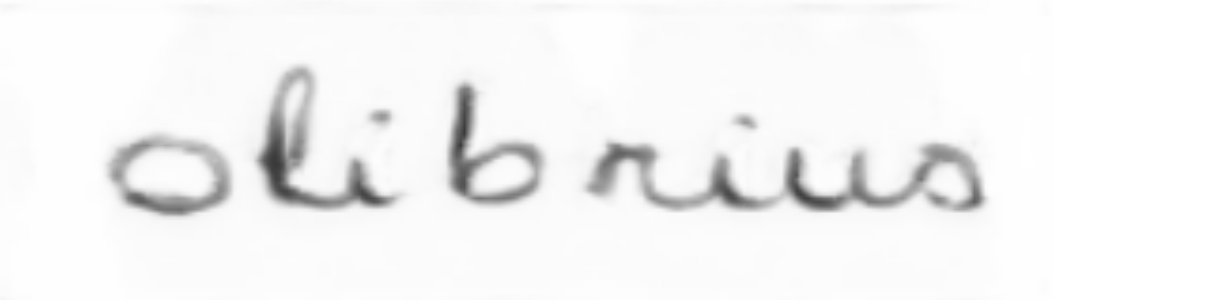} &&
        \includegraphics[height=\myheight,valign=c]{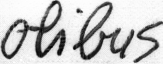} &
        \includegraphics[height=\myheight,valign=c]{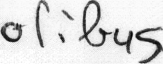} & 
        \includegraphics[height=\myheight,valign=c]{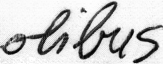} \\ 
        \midrule
        \texttt{"reparer"}&&
        \includegraphics[height=\myheight,valign=c]{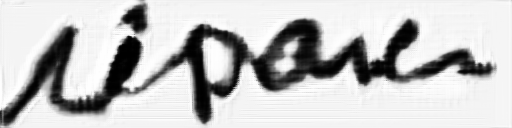} &
        \includegraphics[height=\myheight,valign=c]{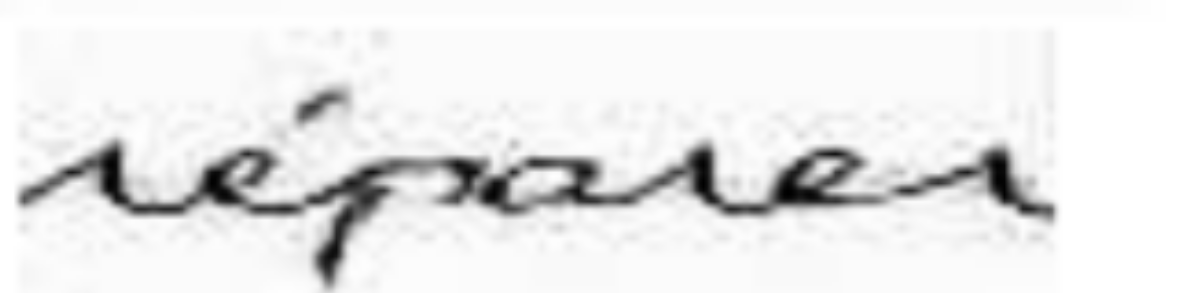} &
        \includegraphics[height=\myheight,valign=c]{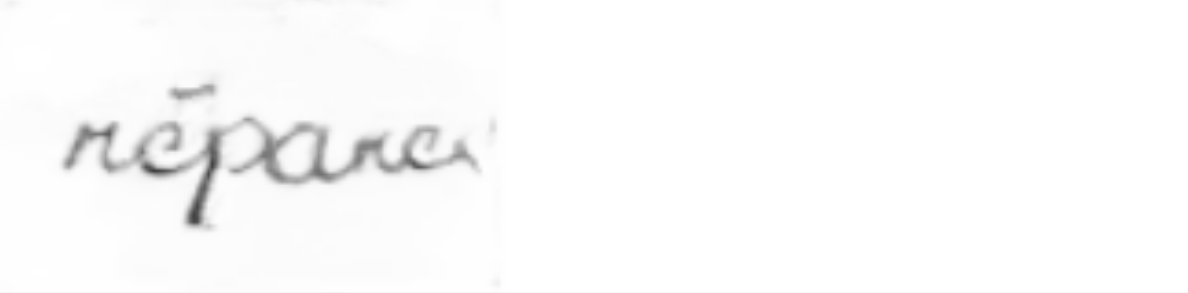} &&
        \includegraphics[height=\myheight,valign=c]{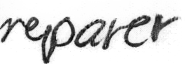} &  
        \includegraphics[height=\myheight,valign=c]{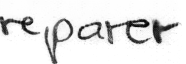} &
        \includegraphics[height=\myheight,valign=c]{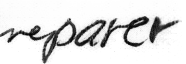} \\ 
        \midrule
        \texttt{"bonjour"}&&
        \includegraphics[height=\myheight,valign=c]{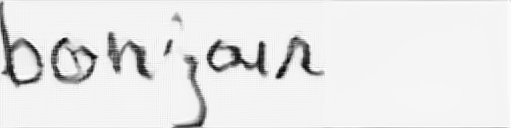} & 
        \includegraphics[height=\myheight,valign=c]{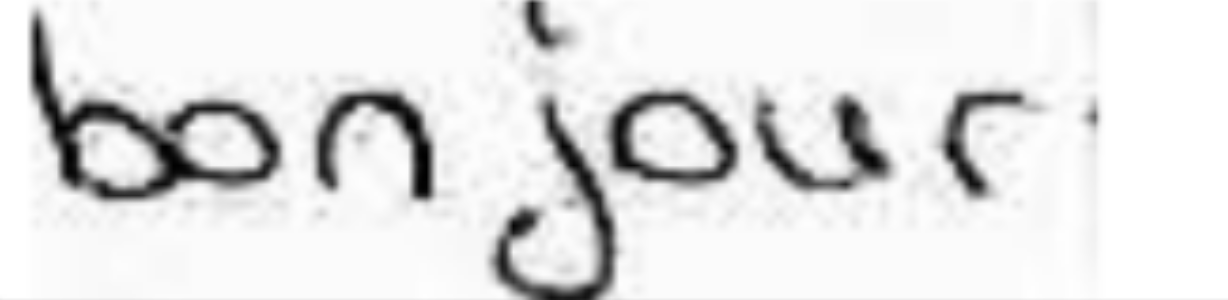} &
        \includegraphics[height=\myheight,valign=c]{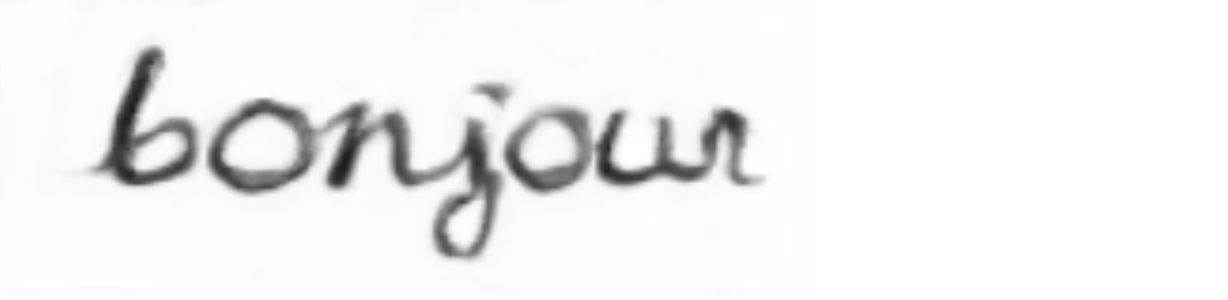} &&
        \includegraphics[height=\myheight,valign=c]{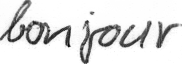} &  
        \includegraphics[height=\myheight,valign=c]{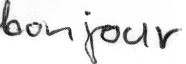} &
        \includegraphics[height=\myheight,valign=c]{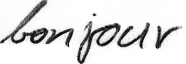} \\  
        \midrule
        \texttt{"famille"}&&
        \includegraphics[height=\myheight,valign=c]{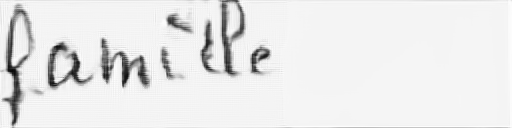} & 
        \includegraphics[height=\myheight,valign=c]{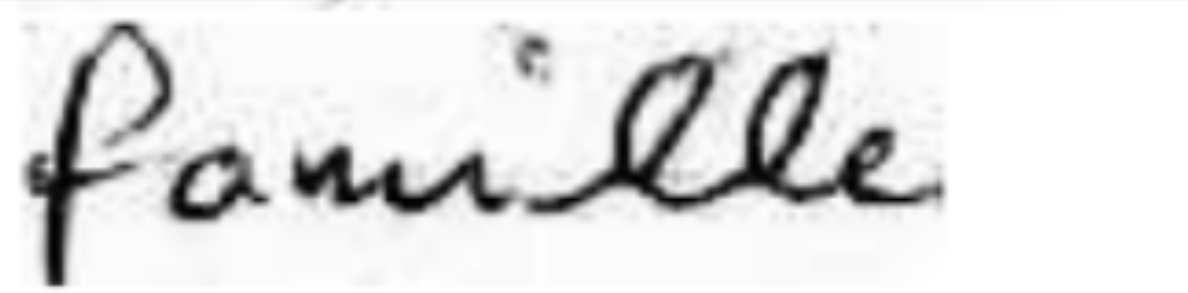} &
        \includegraphics[height=\myheight,valign=c]{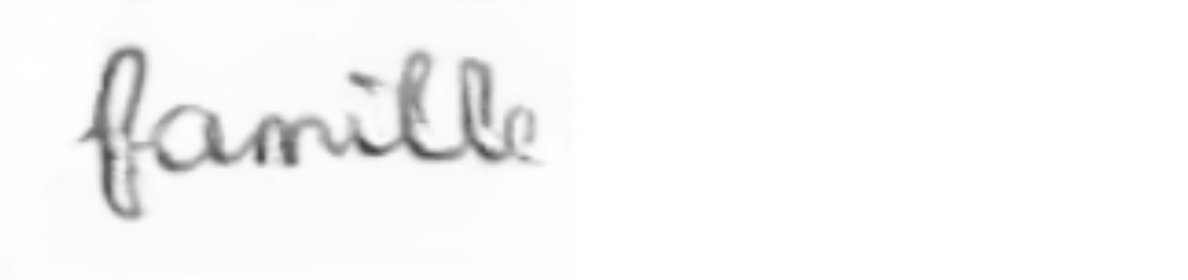} &&
        \includegraphics[height=\myheight,valign=c]{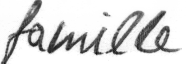} &  
        \includegraphics[height=\myheight,valign=c]{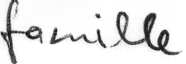} & 
        \includegraphics[height=\myheight,valign=c]{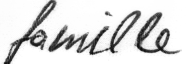} \\
        \midrule
        \texttt{"gorille"}&&
        \includegraphics[height=\myheight,valign=c]{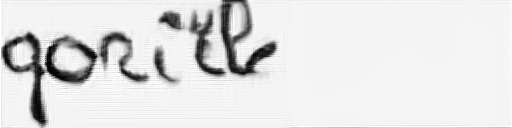} & 
        \includegraphics[height=\myheight,valign=c]{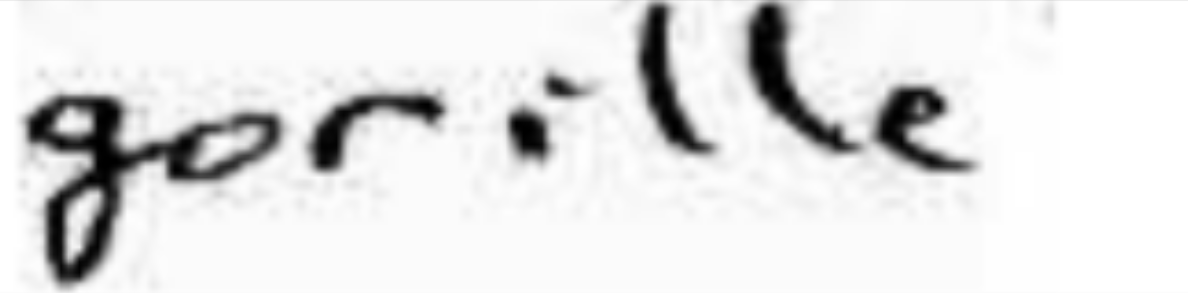} &
        \includegraphics[height=\myheight,valign=c]{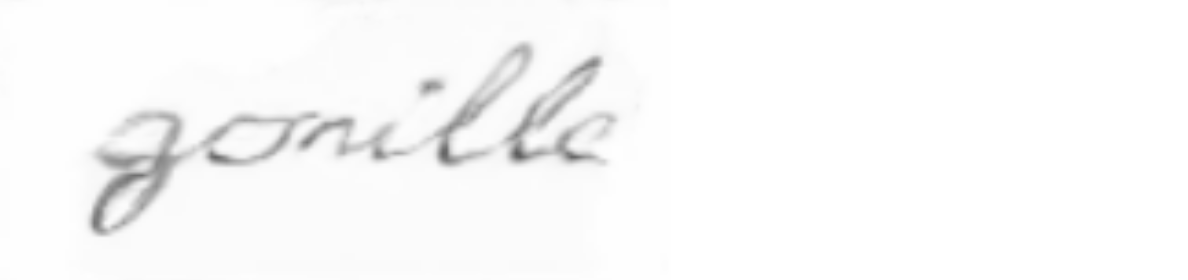} &&
        \includegraphics[height=\myheight,valign=c]{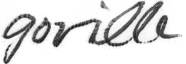} &  
        \includegraphics[height=\myheight,valign=c]{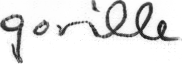} & 
        \includegraphics[height=\myheight,valign=c]{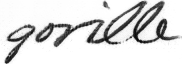} \\  
        \midrule
        \texttt{"malade"}&&
        \includegraphics[height=\myheight,valign=c]{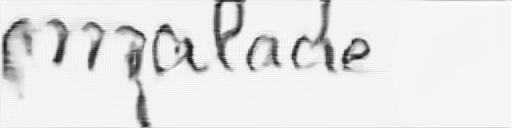} & 
        \includegraphics[height=\myheight,valign=c]{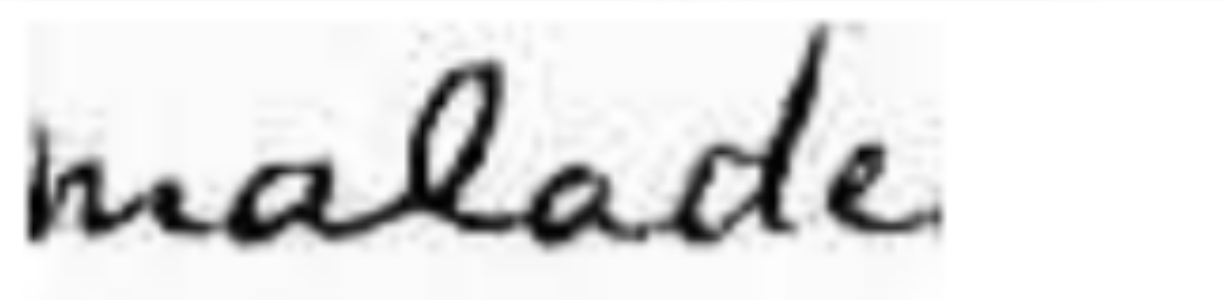} &
        \includegraphics[height=\myheight,valign=c]{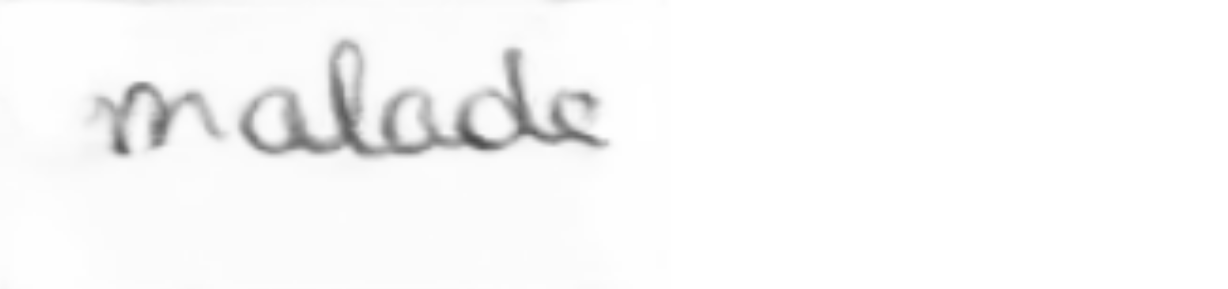} &&
        \includegraphics[height=\myheight,valign=c]{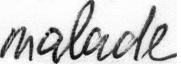} &  
        \includegraphics[height=\myheight,valign=c]{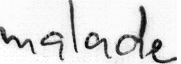} & 
        \includegraphics[height=\myheight,valign=c]{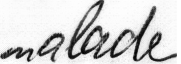} \\  
        \midrule
        \texttt{"certes"}&&
        \includegraphics[height=\myheight,valign=c]{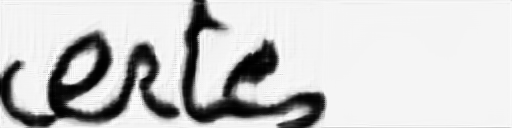} & 
        \includegraphics[height=\myheight,valign=c]{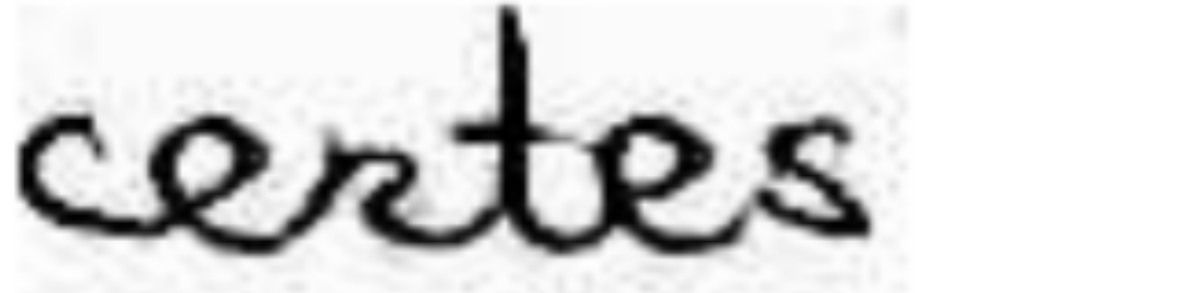} &
        \includegraphics[height=\myheight,valign=c]{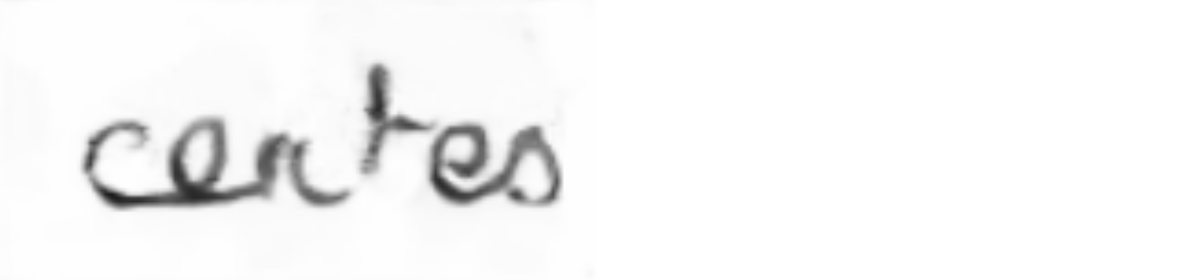} &&
        \includegraphics[height=\myheight,valign=c]{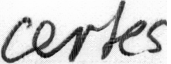} &  
        \includegraphics[height=\myheight,valign=c]{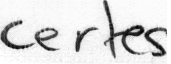} & 
        \includegraphics[height=\myheight,valign=c]{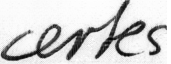} \\ 
        \midrule
        \texttt{"golf"}&&
        \includegraphics[height=\myheight,valign=c]{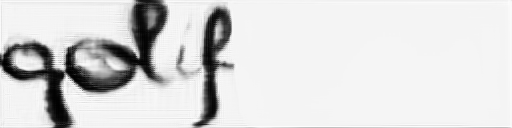} & 
        \includegraphics[height=\myheight,valign=c]{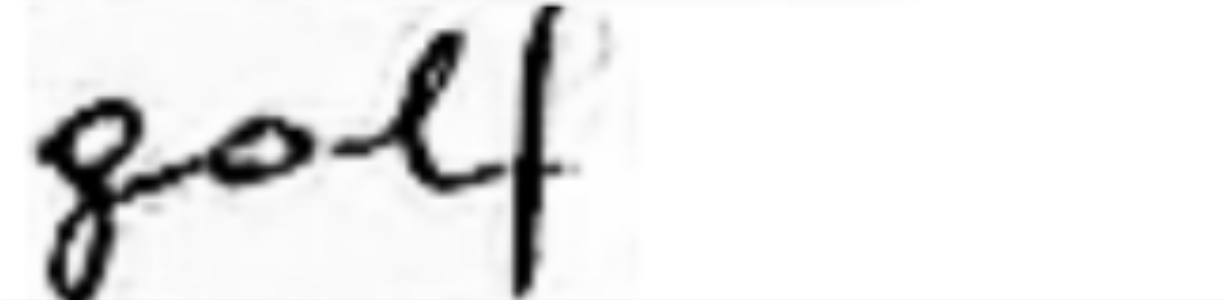} &
        \includegraphics[height=\myheight,valign=c]{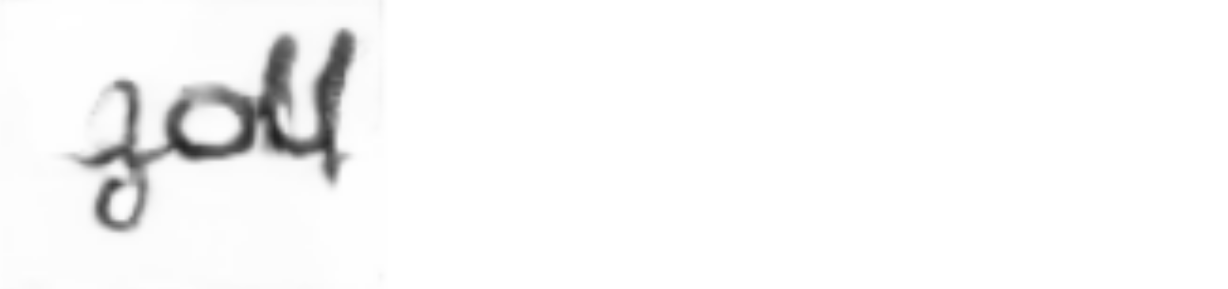} &&
        \includegraphics[height=\myheight,valign=c]{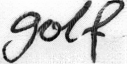} &  
        \includegraphics[height=\myheight,valign=c]{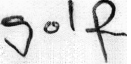} & 
        \includegraphics[height=\myheight,valign=c]{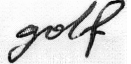} \\ 
        \midrule
        \texttt{"des"}&&
        \includegraphics[height=\myheight,valign=c]{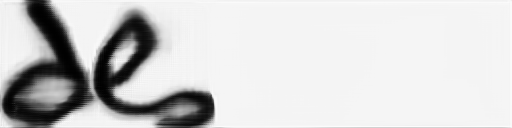} & 
        \includegraphics[height=\myheight,valign=c]{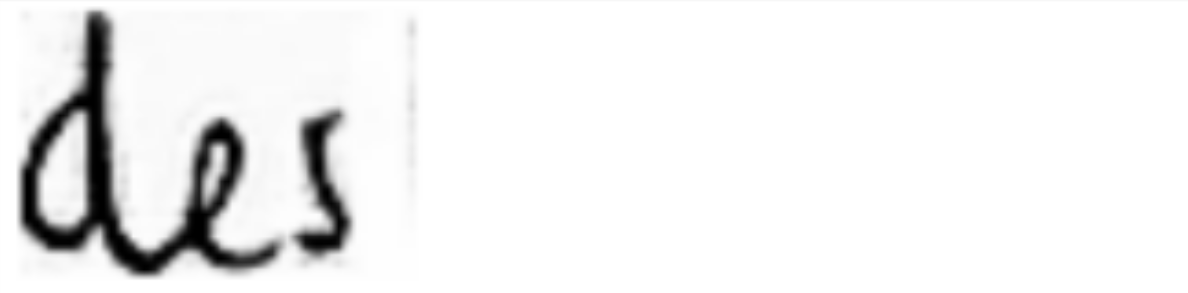} &
        \includegraphics[height=\myheight,valign=c]{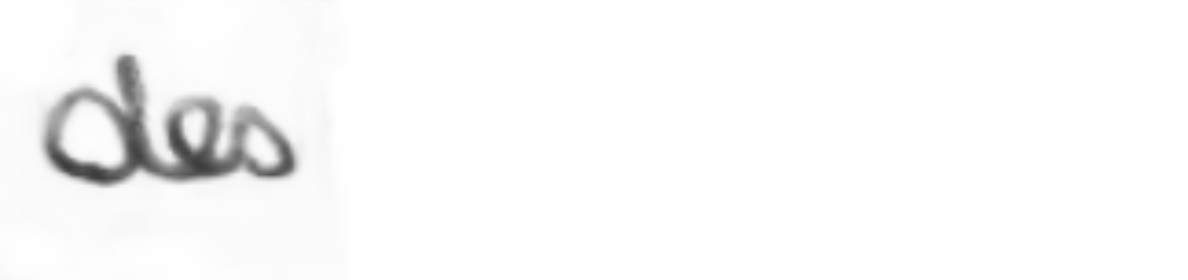} &&
        \includegraphics[height=\myheight,valign=c]{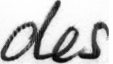} &  
        \includegraphics[height=\myheight,valign=c]{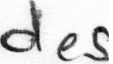} & 
        \includegraphics[height=\myheight,valign=c]{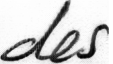} \\  
        \midrule
        \texttt{"ski"}&&
        \includegraphics[height=\myheight,valign=c]{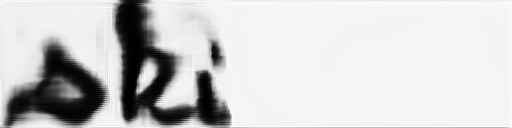} & 
        \includegraphics[height=\myheight,valign=c]{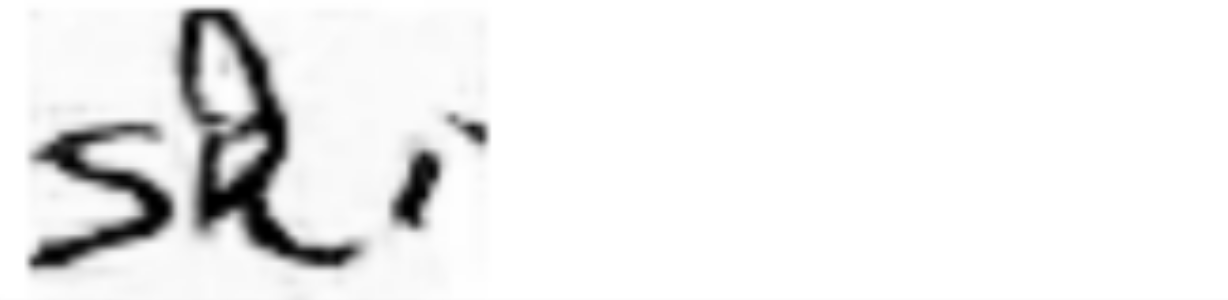} &
        \includegraphics[height=\myheight,valign=c]{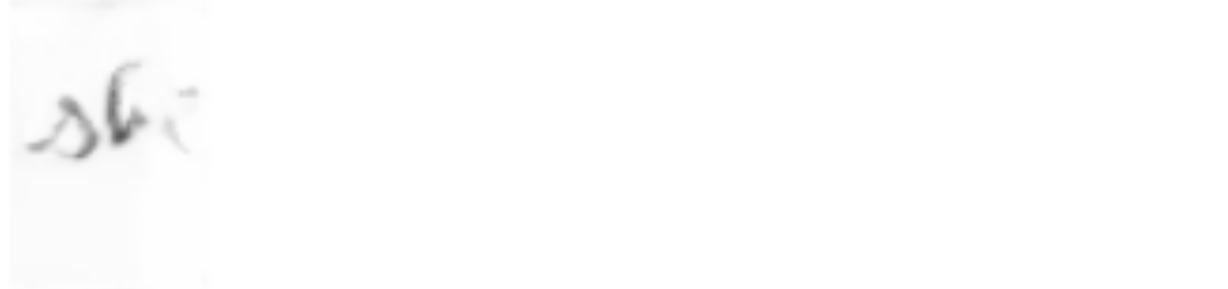} &&
        \includegraphics[height=\myheight,valign=c]{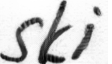} &  
        \includegraphics[height=\myheight,valign=c]{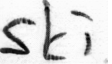} & 
        \includegraphics[height=\myheight,valign=c]{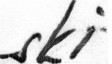} \\  
        \midrule
        \texttt{"le"}&&
        \includegraphics[height=\myheight,valign=c]{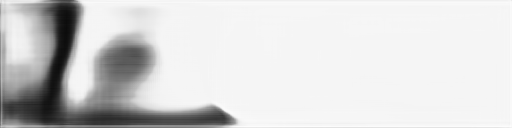} &  
        \includegraphics[height=\myheight,valign=c]{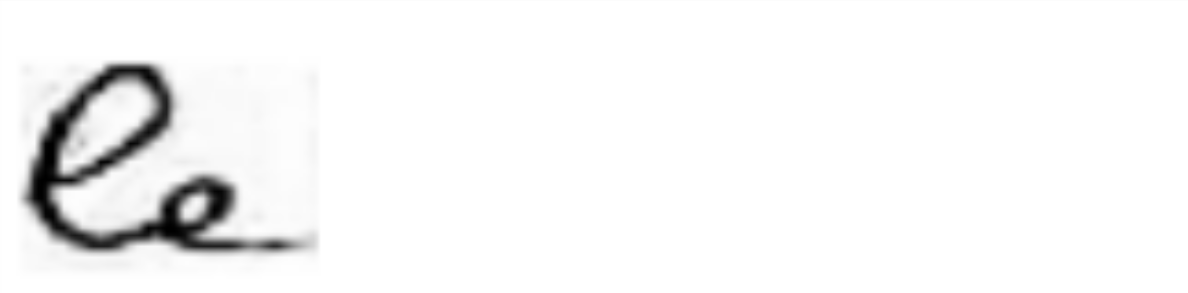} &
        \includegraphics[height=\myheight,valign=c]{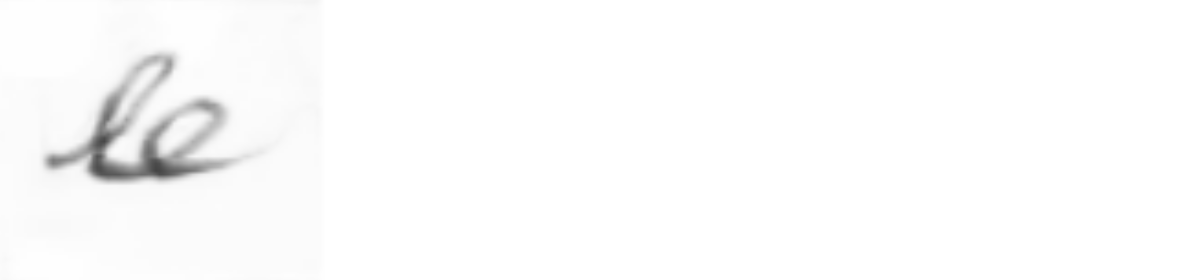} &&
        \includegraphics[height=\myheight,valign=c]{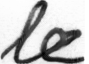} &  
        \includegraphics[height=\myheight,valign=c]{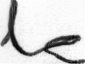} &
        \includegraphics[height=\myheight,valign=c]{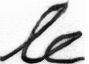} \\  
        \bottomrule
\end{tabular}
%\hspace*{-1cm}
\end{table*}

% \subsection{Comparison with SOTA}
% \begin{figure*}[t!]
%     \centering
%     %\resizebox{\linewidth}{!}{
%     \hspace*{-2cm}
%     \begin{tabular}{ccc}
%     \includegraphics[height=4cm,valign=c]{images/scrabble.png} &
%     \includegraphics[height=4cm,valign=c]{images/bmvc.png} &
%     \missingfigure[figwidth=4cm]{ours}\\
% \end{tabular}
% \hspace*{-2cm}
% %}
%     \caption{Comparison among ScrabbleGAN and BMVCGAN.}
%     \label{fig:comparison_vis}
% \end{figure*}

\subsection{HTR Performance Improvement}
As discussed before, our method has achieved good performance on generating realistic handwritten text-line images with varied styles. These generated data could indeed be used as training data in order to improve the HTR performance at text-line level. For this purpose, we define three settings: first, a conventional supervised learning on the IAM dataset; second, transfer learning from the IAM to the Rimes dataset; and third, a realistic few-shot setting on the Spanish Numbers dataset. To be fairly comparable, we do not use any data augmentation techniques nor pretrained modules.

In all the HTR experiments, we make use of an independently trained handwritten text recognizer, which shares the same architecture with $R$ as detailed in Figure~\ref{fig:transfo} and is trained using the IAM training set at text-line level. If we follow the same experimental setting along the first row (baseline) of Table~\ref{tab:htr} with the jointly trained recognizer, we achieve the CER of $16.73\%$. In contrast, in the first row of Table~\ref{tab:htr}, we can achieve a CER of $10.46\%$ with an independently trained recognizer. From the comparison, we notice that the jointly trained recognizer becomes overfitted during the image generation process. This overfitting effect of the recognizer benefits the image generation, because the overfitted recognizer is sensitive to the data noise, which in return guides the generated image to be cleaner and more readable.

\subsubsection{Enhance the training set}
The most straightforward way to improve the HTR performance is to incorporate extra synthetically generated data to the training set. In Table~\ref{tab:htr}, we evaluate the improvements in different cases. The first row shows the results when using the IAM training set only. To keep the training data balanced between real and generated samples, we generate $8,000$ text-line images based on the style of the IAM images and a lexicon. Concerning the lexicon, we have two choices: WikiText-103 or the groundtruth of IAM training set, shown in the second and third row, respectively. Note that the HTR performance is boosted when adding the $8,000$ synthetically generated samples. Furthermore, the choice of lexicon also matters because the performance is further boosted when using a similar lexicon to the target dataset. Finally, we even apply data augmentation techniques on both the real and generated training samples so that we end up with the best result as shown in the fourth row. Furthermore, our method shows a significant improvement over the performance achieved by ScrabbleGAN~\cite{fogel2020scrabblegan} in comparable settings. Thus, we can conclude that our proposed generative method generates useful samples that are useful to train HTR networks.

\begin{table}[h!]
    \caption{HTR experiments. Results are evaluated on the IAM test set at text-line level. The Error rate reduction is calculated taking the results of the first and last rows.}
    \label{tab:htr}
    \centering
    \begin{tabular}{cccccccc}
        \toprule
        \multirow{2}{*}{Aug.} & \multirow{2}{*}{GAN} & \multirow{2}{*}{Lexicon} &  \multicolumn{2}{c}{ScrabbleGAN~\cite{fogel2020scrabblegan}} &&
        \multicolumn{2}{c}{\textbf{Proposed}}\\
        & & & CER & WER && CER & WER\\
        \midrule
        $-$ & $-$ & IAM & 13.82 & 25.10 && 10.46 & 33.40 \\
        $-$ & \checkmark & WikiText & & && 9.66 & 31.87\\
        $-$ & \checkmark & IAM & & && 9.37 & 30.58\\
        \checkmark & \checkmark & IAM & 13.57 & 23.98 && 8.62 & 26.69\\
        \midrule
        \multicolumn{3}{c}{Error Rate Reduction (\%)} & 1.8 & 4.5 && \textbf{17.6} & \textbf{20.1} \\
        \bottomrule
    \end{tabular}
\end{table}

% \begin{table}[ht!]
%     \caption{HTR experiments. Results are evaluated on IAM test set of text-line level.}
%     \label{tab:htr}
%     \centering
%     \begin{tabular}{ccccc}
%         \toprule
%         Aug. & GAN & Lexicon & CER & WER\\
%         \midrule
%         $-$ & $-$ & IAM & 10.46 & 33.40 \\
%         $-$ & \checkmark & WikiText-103 & 9.66 & 31.87\\
%         $-$ & \checkmark & IAM & 9.37 & 30.58\\
%         \checkmark & \checkmark & IAM & 8.62 & 26.69\\
%         \bottomrule
%     \end{tabular}
% \end{table}

\subsubsection{Transfer learning on a new dataset}
Another useful setting is transfer learning, which consists of transferring a trained recognizer to an unknown dataset. In this case, the source data is the IAM dataset and the target is the Rimes dataset. Both datasets are at text-line level and share some characters in vocabularies such as English letters, space, punctuation marks and numbers. However, the IAM dataset is in English while the Rimes dataset is in French, so some special letters like ``\'e'' or ``\^a'' are exclusive from the Rimes dataset. This scenario may occur in real use cases in which there is a \emph{general} recognizer, which has been properly trained, that is used to recognize a target dataset, containing some exclusive characters. Instead of manually labeling a subset of target data and training the recognizer again, we could provide a faster solution: to generate a set of synthetic samples mimicking the style of target dataset and then fine-tuning on it. In this way, the HTR performance on the target data is boosted to some extent with a manual-free effort, although it can not recognize those special characters.

Table~\ref{tab:transfer_htr} shows the transfer learning results. In the first row, considered as a lower bound, the recognizer is pretrained on the IAM dataset and directly evaluated on the Rimes test set. As an upper bound, we train the recognizer from scratch using the Rimes dataset. Below these two baselines, we show the performance when using the IAM training set and $8,000$ synthetically generated samples using IAM handwriting styles and random text strings from WikiText-103. Secondly, we assume that we have access to images of the Rimes dataset (but not their labels), and we generate $8,000$ synthetic samples that mimic the style of the Rimes dataset while sampling text strings from WikiText-103. We observe that by incorporating these extra synthetically generated samples, the HTR performance for the unlabeled Rimes target dataset is boosted in a transfer learning setting (the CER decreases from 27.3\% down to 18.19\%).

% \begin{table}[ht!]
%     \caption{Transfer learning setting from IAM to Rimes. Results are evaluated on Rimes test set of text-line level.}
%     \label{tab:transfer_htr}
%     \centering
%     \begin{tabular}{cccccc}
%         \toprule
%         Idea & Train set & Style & Lexicon & CER & WER\\
%         \midrule
%         Lower bound & IAM & $-$ & $-$ & 27.30 & 74.57\\
%         w/o target GT & IAM+8k & IAM & WikiText & 20.55 & 63.20\\
%         w/o target GT & IAM+8k & Rimes & WikiText & 18.19 & 54.83 \\
%         Upper bound & Rimes & $-$ & $-$ & 6.45 & 19.56\\
%         \bottomrule
%     \end{tabular}
% \end{table}

% \begin{table}[ht!]
%     \caption{Transfer learning setting from IAM to Rimes. Results are evaluated on Rimes test set of text-line level. Only the second row accesses labeled Rimes data, while the Style column indicates the usage of unlabeled images.}
%     \label{tab:transfer_htr}
%     \centering
%     \begin{tabular}{cccccc}
%         \toprule
%          & Train set & Style & Lexicon & CER & WER\\
%         \midrule
%         Baselines & IAM & $-$ & $-$ & 27.30 & 74.57\\
%          & Rimes & $-$ & $-$ & 6.45 & 19.56\\
%         \midrule
%         Transfer & IAM+8k & IAM & WikiText & 20.55 & 63.20\\
%           & IAM+8k & Rimes & WikiText & 18.19 & 54.83 \\
%         \bottomrule
%     \end{tabular}
% \end{table}

\begin{table}[h!]
    \caption{Transfer learning setting from IAM to Rimes. Results are evaluated on Rimes test set at text-line level. Only the second row has access to labeled Rimes data, while the Adaptation indicates the usage of unlabeled images and external text strings.}
    \label{tab:transfer_htr}
    \centering
    \begin{tabular}{ccccc}
        \toprule
         & Train set & Adaptation & CER & WER\\
        \midrule
        Baselines & IAM  & $-$ & 27.30 & 74.57\\
         & Rimes  & $-$ & 6.45 & 19.56\\
        \midrule
        Transfer & IAM & IAM + WikiText (8K) & 20.55 & 63.20\\
          & IAM & Rimes + WikiText (8K)& 18.19 & 54.83 \\
        \bottomrule
    \end{tabular}
\end{table}

\subsubsection{Few-shot setting on target dataset}

\begin{figure}[h!]
    \centering
    \includegraphics[width=0.7\linewidth]{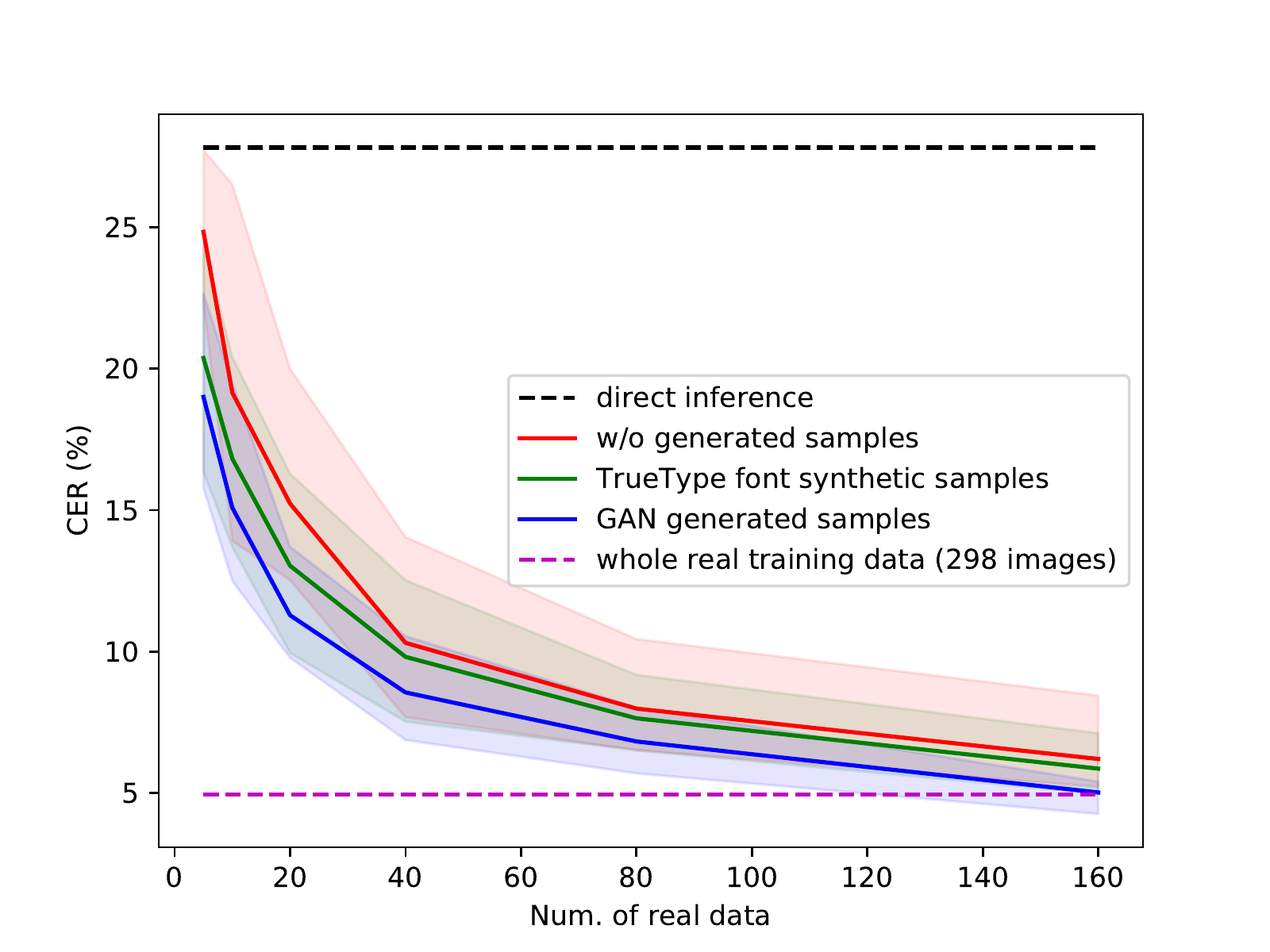}
    \caption{HTR improvement in a real use case on Spanish Number dataset.}
    \label{fig:curve_spanishNum}
\end{figure}

We are also interested in investigating how to make use of the generated images to improve the HTR performance in another realistic scenario: when the target dataset is very small, such as the Spanish Number dataset. Here, we take our baseline method, a recognizer pretrained on IAM dataset, and test it with the test set of Spanish Number data directly, so that we obtain the lower bound with CER 27.82\% as shown as the dashed black line in Figure~\ref{fig:curve_spanishNum}. We hypothesize that we have access to the whole labeled training set of Spanish Number data (298 images), thus we further fine-tune our pretrained recognizer and achieve a CER of 4.94\%, which is the ideal case as shown as the dashed magenta line. 
Then, we select 5, 10, 20, 40, 80, 160 labeled real samples from the Spanish Number training set randomly to carry on the next experiments. Based on our baseline recognizer, we fine-tune on the selected subset of labeled real images to end up with the red curve. Ten individual experiments have been done for each subset of labeled real data, and the data sampling process is also randomly done for every experiment. From the red curve, we can see that the performance is significantly improved with few labeled real samples, while remaining steady when adding more data. 

For comparisons, we carry on experiments with a sequence-to-sequence method that uses synthetic handwritten images based on TrueType fonts~\cite{kang2021candidate}. To avoid a large unbalance between synthetic and real data, we make use of 100, 300, 500, 700 and 900 synthetic data with a specific amount of real subset (indicated as x-axis) to fine-tune the recognizer. We carry on 10 individual experiments with randomly selected synthetic and real subsets, so that we obtain the green curve that behaves better than the red one. The results prove that using extra synthetic data along the training set boosts the HTR performance. However, the handwriting style diversity that the synthetic data provides is very limited to the chosen fonts, so the improvement is also limited.

Since we already have the generative model pretrained on the IAM dataset, we produce synthetic samples based on the unlabeled Spanish Number images and random text strings of WikiText-103. We follow the same experimental setting with the TrueType font based experiments, and generate the blue curve. We observe that our generated samples significantly boost the HTR performance over both the red and green curves. Even more, when using our generated samples with 160 labeled real ones, the recognizer performs better than when using the whole real training set (298 images).

\section{Conclusion and Future Work}
\label{sec:con}
In this work we have presented a generative method to produce realistic and varied artificially rendered samples of handwritten text-line images. With the usage of periodic padding module, the method is able to generate samples of any length disregarding the length of style input. By replacing the Seq2Seq-based recognizer with the Transformer-based one, the ability to generate longer text-line images is obtained. Higher quality results are achieved by training with curriculum learning. Extensive qualitative results have demonstrated the high capacity to generate realistic handwritten text-line images by conditioning the generative process with both visual appearance and textual content information. In addition, our method is able to generate any text-line sample without restriction to any predefined vocabulary, and even work on other languages (except special characters like accents). Also, and once properly trained, the inference can also run in a few-shot setup for the target handwriting style images. 

Furthermore, comprehensive studies on making use of generated samples in both supervised and transfer learning settings have proven that our generated samples can effectively boost the HTR performance with almost no manual effort. Indeed, when comparing to other existing handwritten text generation methods, our method is the one that obtains the most significant HTR improvement (an error reduction of $17.6\%$ in CER and $20.1\%$ in WER).

The method presented in this paper focuses on handwritten data, but in the future, we could further incorporate typed text data. The intuition will be that if the method could generate both cursive handwriting and non-cursive typed text data, the visual style transfer from any random handwritten images to unified typed text samples may achieve a good performance, which could drastically improve the HTR performance nowadays.

%In the future, we plan to collect a large available handwritten text-line dataset with different languages including English, French, Spanish and German, so that a general generative approach could produce text strings with special accent characters.  

\ifCLASSOPTIONcompsoc
  \section*{Acknowledgments}
\else
  \section*{Acknowledgment}
\fi
%\section*{Acknowledgments}
This work has been partially supported by the grant 140/09421059 from Shantou University, the Spanish project RTI2018-095645-B-C21, the grant 2016-DI-087 from the Secretaria d'Universitats i Recerca del Departament d'Economia i Coneixement de la Generalitat de Catalunya, the Ramon y Cajal Fellowship RYC-2014-16831 and the CERCA Program/ Generalitat de Catalunya.

% This work has been partially supported by the Spanish project RTI2018-095645-B-C21, the grant 2016-DI-087 from the Secretaria d'Universitats i Recerca del Departament d'Economia i Coneixement de la Generalitat de Catalunya, the grant FPU15/06264 from the Spanish Ministerio de Educaci\'{o}n, Cultura y Deporte, the Ramon y Cajal Fellowship RYC-2014-16831 and the CERCA Program/ Generalitat de Catalunya. We gratefully acknowledge the support of NVIDIA Corporation with the donation of the Titan Xp GPU used for this research.

\ifCLASSOPTIONcaptionsoff
  \newpage
\fi

\bibliographystyle{IEEEtran}
\bibliography{bib.bib}

\end{document}